\documentclass[final]{cvpr}

\usepackage{times}
\usepackage{epsfig}
\usepackage{graphicx}
\usepackage{amsmath}
\usepackage{amssymb}
\usepackage{multirow}
\usepackage{animate}
\usepackage{subfigure}

\usepackage[pagebackref=true,breaklinks=true,colorlinks,bookmarks=false]{hyperref}

\def\cvprPaperID{4280} 
\def\confYear{CVPR 2021}
\begin{document}

\title{From Shadow Generation to Shadow Removal}
\author{Zhihao Liu$^{1}$, Hui Yin$^{1,*}$, Xinyi Wu$^{2}$, Zhenyao Wu$^{2}$, Yang Mi$^{3}$, Song Wang$^{2,}$\thanks{Co-corresponding authors. Code is available at \url{https://github.com/hhqweasd/G2R-ShadowNet}.}\\{\normalsize $^1$Beijing Jiaotong University, China\qquad $^2$University of South Carolina, USA\qquad $^3$China Agriculture University, China}\\{\tt\small\{16120394,hyin\}@bjtu.edu.cn,\{xinyiw,zhenyao\}@email.sc.edu,miy@cau.edu.cn,songwang@cec.sc.edu}
}

\maketitle
\pagestyle{empty}
\thispagestyle{empty}

\begin{abstract}
 Shadow removal is a computer-vision task that aims to restore the image content in shadow regions. While almost all recent shadow-removal methods require shadow-free images for training, in ECCV 2020 Le and Samaras introduces an innovative approach without this requirement by cropping patches with and without shadows from shadow images as training samples. However, it is still laborious and time-consuming to construct a large amount of such unpaired patches. In this paper, we propose a new G2R-ShadowNet which leverages shadow generation for weakly-supervised shadow removal by only using a set of shadow images and their corresponding shadow masks for training. The proposed G2R-ShadowNet consists of three sub-networks for shadow generation, shadow removal and refinement, respectively and they are jointly trained in an end-to-end fashion. In particular, the shadow generation sub-net stylises non-shadow regions to be shadow ones, leading to paired data for training the shadow-removal sub-net. Extensive experiments on the ISTD dataset and the Video Shadow Removal dataset show that the proposed G2R-ShadowNet achieves competitive performances against the current state of the arts and outperforms Le and Samaras’ patch-based shadow-removal method.
\end{abstract}
	
\section{Introduction}

Shadows are areas of darkness in a scene where the light is fully or partially occluded. Shadows are very common in natural images and might bring challenges to many existing computer vision tasks~\cite{nadimi2004physical,jung2009efficient,le2018adnet,le2019weakly}. Shadow removal by restoring the image information in shadow regions have been a long studied research problem~\cite{cucchiara2003detecting,le2016geodesic,su2016shadow,le2017co,zhang2018improving,muller2019brightness} and has been shown to be beneficial to improve the performance in various tasks.

	\begin{figure}[htbp]\small
		\centering
		\includegraphics[width=1\linewidth]{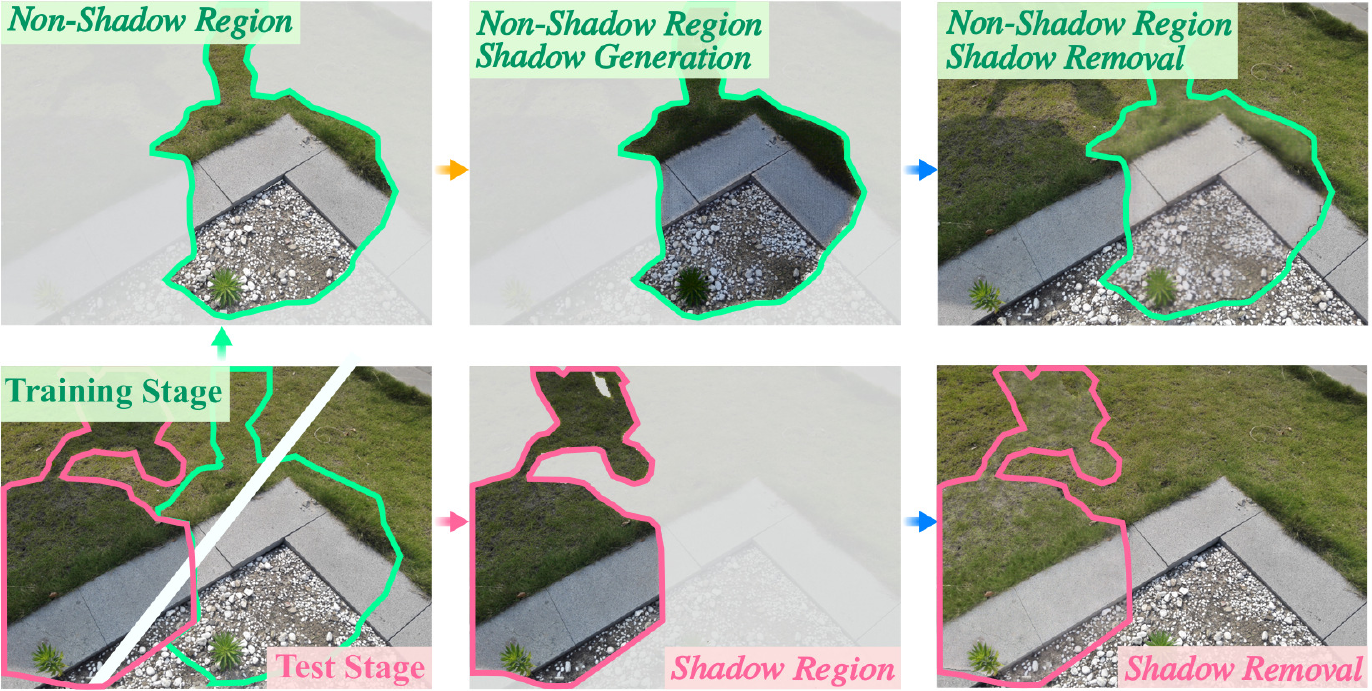}
		\caption{An illustration of our basic idea of incorporating shadow generation for learning shadow removal by using only shadow images. The pixels located in the pink and green boundaries of the input shadow image form the shadow and random non-shadow regions, respectively. The pink and green arrows stand for the masking operation that only preserves the region with the value of 1 on the mask, while the orange and blue arrows represent the process of shadow generation and shadow removal, respectively.}
		\vspace{-13pt}
		\label{fig:1}
	\end{figure}
	
Recently, with the use of the convolutional neural networks (CNNs), many learning based shadow removal approaches~\cite{qu2017deshadownet,wang2018stacked,hu2019direction,hu2019mask,Le2019Shadow,liu2021shadow,lin2020bedsr,le2020from} have been proposed, resulting in significantly better performance than the traditional ones~\cite{finlayson2005removal,guo2012paired,khan2015automatic,zhang2015shadow}. 
For most of them~\cite{qu2017deshadownet,wang2018stacked,hu2019direction,Le2019Shadow} a set of paired shadow images and their corresponding shadow-free version are used to train the network in a fully-supervised manner. However, it is difficult to capture and collect such paired image data in uncontrolled natural environment due to the illumination change from time to time. If we capture such data pairs in a controlled lab environment, limited scenarios might also weaken the generalisation ability of the trained model.

To address these problems, recent researches~\cite{hu2019mask,liu2021shadow} start to explore unsupervised methods for shadow removal by using unpaired shadow and shadow-free images. However, these unsupervised methods might introduce huge domain gap between the shadow and shadow-free images in the training set. In addition, in practice, it is still difficult to capture a large set of shadow-free images with good variety. To handle this problem, in their ECCV20 paper~\cite{le2020from}, Le and Samaras introduces a novel unsupervised shadow removal method using only shadow images. More specifically, they leverage the fact that a shadow image usually contains both shadow and non-shadow regions. This way, a set of shadow and shadow-free patches can be cropped to construct unpaired data for network training. By cropping unpaired patches from the same images, their domain gap can also be well controlled. 

Although the patch cropping in~\cite{le2020from} only requires the shadow masks that can be obtained by using an existing shadow detection method~\cite{zhu2018bidirectional,le2018adnet,zheng2019distraction,hu2019direction,wang2020instance}, it involves a careful design of cropping-window size, a set of strict physics-based constraints, and heavy computational load~\cite{le2020from}. In this paper, we propose a new approach to address these problems by incorporating a shadow generation module, while keeping the desirable property of using only shadow images, as illustrated in Fig.\ref{fig:1}.
	
Specifically, we propose a new G2R-ShadowNet that consists of three sub-networks for shadow generation, shadow removal and refinement, respectively. Given an input shadow image, the shadow-generation sub-net generates pseudo shadows for each shadow-free region and such pseudo shadows are then paired with the corresponding original shadow-free region to form the training data. After that, these constructed pair data are used to train the shadow removal sub-net to remove the generated shadows. Finally, the shadow-removal results are refined by leveraging the context information such that their colour and illumination are consistent with their surrounding areas. We conduct extensive experiments on the ISTD dataset and the Video Shadow Removal dataset to demonstrate the effectiveness of our proposed method.
	
The main contributions of this work are as follows:
\begin{itemize}
\vspace{-4pt}
\item We tackle the shadow removal task from a novel perspective of constructing paired shadow and non-shadow data using only the shadow images and the corresponding shadow masks. 
\vspace{-4pt}
\item We develop G2R-ShadowNet, a novel shadow-removal network, which consists of three sub-nets for shadow generation, shadow removal, and refinement, respectively. G2R-ShadowNet is weakly-supervisedly trained in an end-to-end fashion.
\vspace{-4pt}
\item We conduct extensive experiments on two public datasets and show that the proposed G2R-ShadowNet achieves competitive performances against the current state of the arts and outperforms Le and Samaras’ patch-based shadow-removal method.
\end{itemize}

\section{Related Work}


\subsection{Shadow generation}
	
Our proposed G2R-ShadowNet contains a shadow generator which employs the generative adversarial networks (GAN)~\cite{goodfellow2014generative} for generating shadows on shadow-free regions. 
GAN-based shadow generation has been studied by many researchers.  
Zhang \etal~\cite{zhang2019shadowgan} proposed a GAN to synthesise shadows for virtual objects that are inserted into images and train the network with shadow masks and paired shadow/shadow-free images. Similarly, Liu \etal~\cite{liu2020arshadowgan} developed a ARShadowGAN for augmented reality in single light scenes, which exploits attention mechanism to model the mapping relationship between the shadow of the virtual objects and the real-world environment. These two works~\cite{zhang2019shadowgan,liu2020arshadowgan} rely on fully-supervised training which requires the shadow and shadow-free images as well as the shadow masks.  
In ~\cite{hu2019mask,liu2021shadow}, a mapping is learned between the unpaired shadow and shadow-free images for shadow removal, in which shadow generators are trained to match the distributions of the generated shadows and the real shadows based on unpaired images.	
However, the shadow-free images are the prerequisites for all the above methods to generate the shadow images. In contrast, in this paper we sample both shadow and shadow-free regions only from shadow images, which is also different from other methods that need images from target domain for their respective applications~\cite{wen2019single,brooks2019learning} or generate data on the whole image~\cite{xu2020noisy}.
	
\subsection{Shadow removal}
Traditional approaches remove shadows according to image gradients~\cite{finlayson2005removal,gryka2015learning}, illumination information~\cite{shor2008shadow,yang2012shadowremoval,xiao2013fast,zhang2015shadow}, and region properties~\cite{guo2012paired,vicente2017leave}.  Recently, supervised learning based methods trained with large-scale paired datasets boost the shadow-removal performance significantly~\cite{qu2017deshadownet,wang2018stacked,hu2019direction,ding2019argan,Le2019Shadow,lin2020bedsr}. However, as mentioned above, paired shadow and shadow-free images are difficult to obtain in practice. To get rid of the dependence on paired data, Hu \etal~\cite{hu2019mask} proposed a Mask-ShadowGAN framework based on the CycleGAN~\cite{zhu2017unpaired}, which leverages unpaired data to learn the adaptation from the shadow-free domain to the shadow domain and vice versa. Liu \etal~\cite{liu2021shadow} later developed a LG-ShadowNet framework to improve the Mask-ShadowGAN~\cite{hu2019mask} by introducing a lightness-guided strategy, which uses the learned lightness features to guide the learning of shadow removal.

However, all these methods still need shadow-free images for training which requires very strict acquisition conditions and may introduce huge gap between the source (non-shadow) domain and the target (shadow) domain. In this paper, the shadow and non-shadow regions are sampled from the same shadow image and therefore, the distribution difference between the two domains is much smaller. In addition, we apply the adversarial training only in the shadow-generation sub-net, while Mask-ShadowGAN and LG-ShadowNet apply it in both shadow-generation and shadow-removal networks, which makes the whole framework more difficult to converge.

    \begin{figure*}[htbp]\small
		\centering
		\includegraphics[width=1\linewidth]{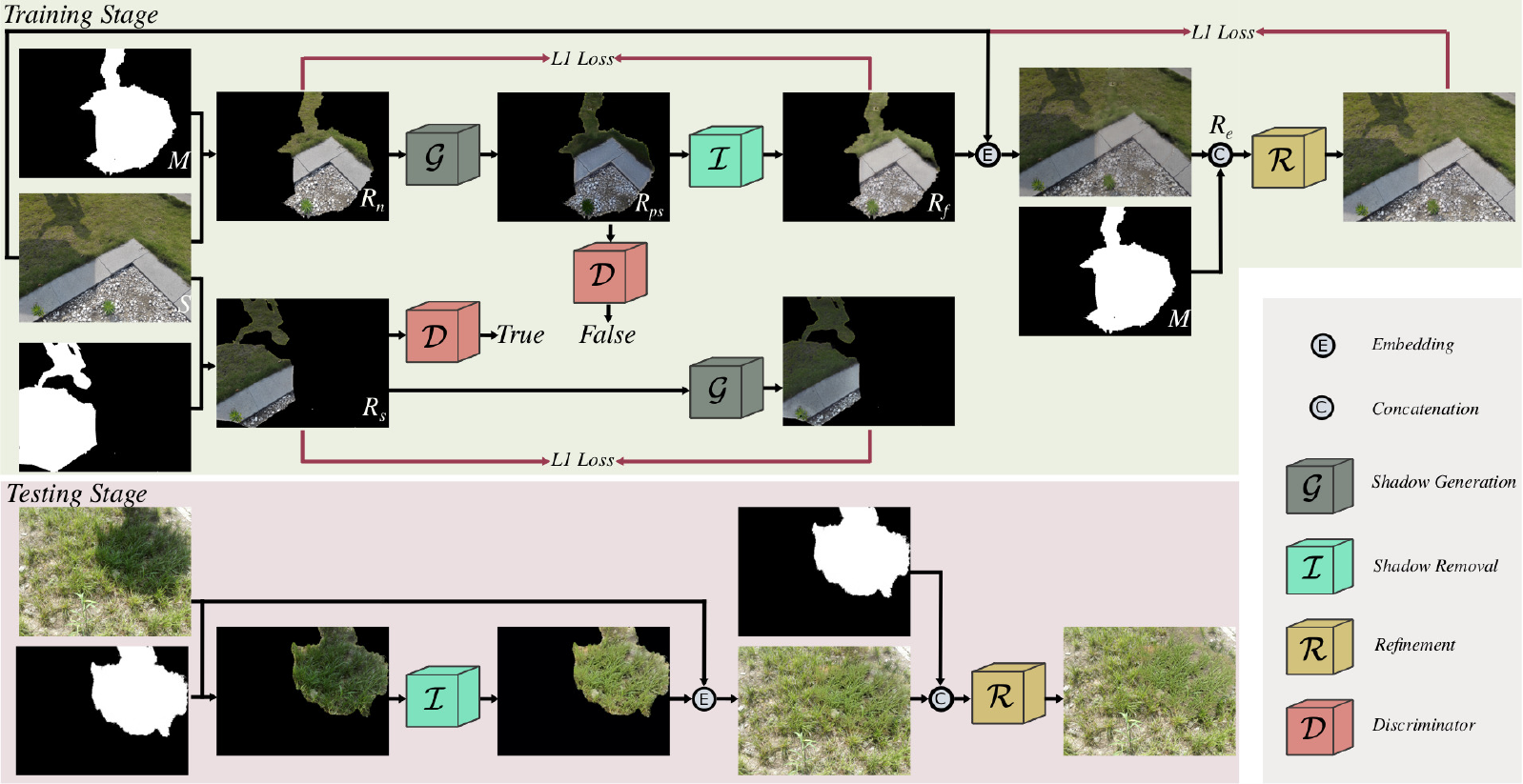}
		\caption{The network architecture of our proposed G2R-ShadowNet. It takes in a shadow image and its corresponding shadow mask to produce the shadow-free result in the shadow regions. The training stage involves all the three sub-nets of shadow generation, shadow removal and refinement, while the testing stage does not perform shadow generation.}
		\label{fig:model}
		\vspace{-5pt}
    \end{figure*}
	
As mentioned above, most related to our work is~\cite{le2020from}, where unpaired data in form of patches are cropped from the same shadow image according to the shadow mask. Therefore, the constructed data show small domain gap. However, this patch-based method suffers from a heavy computational load because of the repetitive cropping of a small step size. In addition, the strict physics-based constraints used in this method limits the shadow types that can be handled. Our proposed G2R-ShadowNet can address these issues by constructing paired data from the same shadow image using the shadow mask without repetitive cropping.

\section{Methodology}

In this section, we elaborate on the overall network architecture of our proposed G2R-ShadowNet, which includes three modules: shadow generation, shadow removal, and refinement. All the three parts are jointly trained in an end-to-end fashion, as illustrated in Fig.~\ref{fig:model}.
	
\subsection{Shadow generation sub-net}\label{31}
	
The shadow generation sub-net is used to construct our training data, i.e., paired shadow and non-shadow regions, for the shadow removal network. Specifically, we first crop the shadow region $R_s$ from the input image $S$ by applying the corresponding shadow mask: the remaining area of the image is setting to 0 automatically. Then we randomly pick another shadow mask $M$ from the masks of the training set and apply it on the shadow-free region of $S$ to obtain a non-shadow region $R_n$, whose area approximates the area of the shadow region $R_s$ by the constraint
	\begin{equation}
	\begin{aligned}
        \mathtt{Area}(R_n)/\mathtt{Area}(R_s)\in(1-\alpha,1+\alpha),
	\end{aligned}
	\end{equation}
where $\mathtt{Area}(\cdot)$ computes the area of the given region and $\alpha$ is a tolerance value which is set to 0.2 in our experiments. Note that this constraint will be discarded in the cases where $R_s$ covers more than half of the whole image and we randomly select a non-shadow region for masking and shadow generation.
	
Using the above operations, we construct many pairs of unaligned data from both shadow and non-shadow domains, which are used to train our shadow generator $\mathcal{G}$ to generate pseudo shadow $R_{ps}$ on the non-shadow region of $S$ via adversarial training. A discriminator $\mathcal{D}$ is employed to distinguish the pseudo shadow $R_{ps}$ from a randomly sampled real shadow to help train $\mathcal{G}$ and ensure the data distribution similarity between the two domains.
	
The architecture of $\mathcal{G}$ mainly follows the generator proposed by Hu \etal~\cite{hu2019mask}. It consists of three convolutional layers with a stride of 2 to decrease the resolution of the input image, followed by nine residual blocks to extract features, and ends with three deconvolutional layers to generate the output with the same resolution as $S$. Besides, the instance normalisation~\cite{ulyanov2016instance} is applied after each convolutional operation. For the architecture of $\mathcal{D}$, we directly employ the one proposed in PatchGAN~\cite{isola2017image}. All the inputs and outputs of $\mathcal{G}$ and the inputs of $\mathcal{D}$ are 3-channel images in LAB colour space.
	
The objective functions to train the shadow generator $\mathcal{G}$ and the discriminator $\mathcal{D}$ are defined as:
	\begin{equation}
	\begin{aligned}
	L_{Gen}(\mathcal{G})=&\frac{1}{2}\mathbb{E}_{R_n \sim p(R_n)}\big[(\mathcal{D}(\mathcal{G}(R_n))-1)^2\big],\\
	\end{aligned}
	\end{equation}
	\begin{equation}
	\begin{aligned}
	L_{Dis}(\mathcal{D})=&\frac{1}{2}\mathbb{E}_{R_n \sim p(R_n)}\big[(\mathcal{D}(\mathcal{G}(R_n)))^2\big]\\
	+&\frac{1}{2}\mathbb{E}_{R_s \sim p(R_s)}\big[(\mathcal{D}(R_s)-1)^2\big].
	\end{aligned}
	\end{equation}
The combined loss function for the adversarial training is:
	\begin{equation}
	\begin{aligned}
	L_{GAN}=&L_{Gen}(\mathcal{G})+L_{Dis}(\mathcal{D}).\\
	\end{aligned}
	\end{equation}
	
In addition, to ensure that the shadow generation sub-net produces high quality synthetic shadows, real shadow $R_s$ is also fed to $\mathcal{G}$, and we apply the identical loss~\cite{taigman2016unsupervised} to encourage $\mathcal{G}$ to generate the same shadow as the input $R_s$, which is defined as:
	\begin{equation}
	\begin{aligned}
	L_{iden}&(\mathcal{G})=\mathbb{E}_{R_s\sim p(\mathcal{R}_s)}\big[\left \|\mathcal{G}(R_s),R_s\right \|_1\big],
	\end{aligned}
	\end{equation}
	where $\left \|,\right \|_1$ represents $L_1$ loss.
	
\subsection{Shadow removal sub-net}

We use the pairs of $R_{ps}$ and $R_{n}$ as the inputs of our shadow removal sub-net for learning to remove the pseudo shadows that are generated by the generator $\mathcal{G}$. Specifically, the shadow removal sub-net $\mathcal{I}$ has the same structure as the generator $\mathcal{G}$, and it takes the output of $\mathcal{G}$, i.e., $R_{ps}$, as the input to produce a shadow-free result $R_f$ that shares the same content as $R_{ps}$. The loss function to train $\mathcal{I}$ is defined as:
	\begin{equation}
	\begin{aligned}
	L_{rem}(\mathcal{G},\mathcal{I})=&\mathbb{E}_{R_{ps}\sim p(\mathcal{R}_{ps})}\big[\left \|\mathcal{I}(R_{ps}),R_n\right \|_1\big]\\
	=&\mathbb{E}_{R_{n}\sim p(\mathcal{R}_{n})}\big[\left \|\mathcal{I}(\mathcal{G}(R_{n})),R_n\right \|_1\big].
	\end{aligned}
	\end{equation}
Note that the gradient computed by this loss will be propagated back to the shadow generator $\mathcal{G}$ through $R_{ps}$ and therefore, it can be regarded as a cycle loss~\cite{zhu2017unpaired} for training both $\mathcal{G}$ and $\mathcal{I}$.

\subsection{Refinement sub-net}\label{33}
	
The output $R_f$ from the shadow removal sub-net is then embedded into the input image $S$ to obtain $R_e$ which is formulated as:
	\begin{equation}
	R_{e}=\left \langle R_{f}+S-R_n,M \right \rangle,
	\label{equ:replace}
	\end{equation}
where $\left \langle \cdot , \cdot \right \rangle$ denotes the concatenation operation, and $M$ is the shadow mask used in previous parts which covers the region to be processed. The obtained $R_e$ is a 4-channel tensor including 3 channels for the image and 1 channel for the mask. However, the colour of $R_f$ might not be fully consistent with that of the other regions of image $S$. To address this problem, we further develop a refinement sub-net $\mathcal{R}$ to refine $R_e$ by exploiting context information of the shadow region over the original whole shadow image.
	
Specifically, our refinement network $\mathcal{R}$ takes in $R_e$ as input and outputs a refined image $R_r$. The network $\mathcal{R}$ shares the same structure as both $\mathcal{I}$ and $\mathcal{G}$ except for the number of the input channels. A per-pixel loss between the refined image $R_r$ and the input $S$ is computed to train the refinement network, which is defined as:
	\begin{equation}
	\begin{aligned}
	L_{full}&(\mathcal{G},\mathcal{I},\mathcal{R})\\
	=&\mathbb{E}_{R_{f}\sim p(\mathcal{R}_{f})}\big[\left \|\mathcal{R}(R_{e}),S\right \|_1\big].\\
\end{aligned}
	\end{equation}
This loss function is calculated according to the context information of the shadow region across the whole shadow image and the computed gradient will be propagated back to $\mathcal{G}$ and $\mathcal{I}$.
	
To further emphasise the content of the output in the same region as $R_n$ to be the same as that part in $S$, we apply the following loss function: 
	\begin{equation}
	\begin{aligned}
	L_{area}&(\mathcal{G},\mathcal{I},\mathcal{R})\\
	=&\mathbb{E}_{R_{f}\sim p(\mathcal{R}_{f})}\big[\sum_n \psi (M)|\mathcal{R}(R_{e})-S|\big],\\
	\end{aligned}
	\label{equ:dilation}
	\end{equation}
where $n$ is the number of pixels in the input image $S$. $\psi$ denotes the image dilation function with a kernel size of $\tau$, which produces a dilated mask to guide the model to pay more attention to the adjacent area of $R_n$. 
	
\subsection{Loss function}

By combining all the loss functions proposed for the above three sub-nets, the total loss $\mathcal{L}$ for training the shadow generator $\mathcal{G}$, the shadow removal sub-net $\mathcal{I}$ and the refinement sub-net $\mathcal{R}$ is defined as
	\begin{equation}
	\begin{aligned}
	\mathcal{L}=&\omega_1 L_{GAN}+\omega_2 L_{iden}\\
	&+\omega_3 L_{rem}+\omega_4 L_{full}+\omega_5 L_{area},
	\label{equ:loss}
	\end{aligned}
	\end{equation}
where $\omega_1$, $\omega_2$, $\omega_3$, $\omega_4$, and $\omega_5$ are the weights to balance different loss terms and are set to $1.0$, $5.0$, $1.0$, $1.0$, and $1.0$, respectively in our experiments.

\section{Experiments}
	
\subsection{Datasets and evaluation metrics}
\noindent{\bf ISTD~\cite{wang2018stacked,Le2019Shadow}}\quad The ISTD dataset is proposed for both shadow detection and shadow removal and the data are collected under various illumination conditions with different shadow shapes. In total, it contains 1,870 triplets of shadow, shadow mask and shadow-free images with a resolution of $480 \times 640$, where 1,330 triplets for training the rest 540 for testing. Following~\cite{Le2019Shadow}, we apply the adjusted testing set with reduced illumination difference between the shadow and shadow-free images in the original dataset. In training, our model uses the shadow images and the corresponding ground-truth shadow masks. In testing, we employ the shadow detector proposed by Zhu \etal~\cite{zhu2018bidirectional} to obtain the shadow mask of the shadow images in the test set. The shadow detector is trained on both the training set of ISTD and the SBU~\cite{vicente2016large} datasets, and it achieves 2.4 Balance Error Rate on the testing set of ISTD when evaluated using the ground-truth shadow masks provided by ISTD.

\vspace{3pt}
\noindent{\bf Video Shadow Removal Dataset~\cite{le2020from}}\quad The Video Shadow Removal dataset contains 8 videos whose contents are static scenes without moving objects. This dataset also provides a corresponding $V_{max}$ image for each video and a moving-shadow mask for each frame. The $V_{max}$ image is obtained by taking the maximum intensity value at each pixel location across the whole video, which is regarded as the shadow-free ground truth of the video. The moving-shadow mask covers the pixels appearing in both the shadow and shadow-free regions of the video, which represents the region for evaluation. We follow the setting in the official code of~\cite{le2020from} using a threshold of 80 to obtain the moving-shadow mask. For this data, we apply the shadow detector~\cite{zhu2018bidirectional} that is trained only on the SBU dataset to generate the shadow masks for our experiments.

\vspace{3pt}
\noindent{\bf Evaluation metrics}\quad For all experiments, we use the Root-Mean-Square Error (RMSE), Peak Signal-to-Noise Ratio (PSNR) and Structural Similarity (SSIM) as the evaluation metrics. Following~\cite{wang2018stacked,hu2019mask,Le2019Shadow,liu2021shadow,le2020from}, we compute the RMSE between the produced shadow-free image and the ground-truth images in the LAB colour space. While some recent work~\cite{le2020from} computes RMSE at each pixel and then averages the score over all the pixels, we compute RMSE \textit{on each image} and then average the score over all images/frames for both the ISTD dataset and the video dataset. Our computed RMSE emphasise more the quality of each image on shadow and non-shadow regions and is more consistent with other metrics such as PSNR and SSIM. We also compute PSNR and SSIM scores in the RGB colour space to evaluate our method. For RMSE, the lower the better while for PSNR and SSIM, the higher the better.
	
\subsection{Experimental settings}
We implement our proposed G2R-ShadowNet using PyTorch with a single NVIDIA GeForce GTX 2080ti GPU. We initialise our model using a Gaussian distribution with a mean of $0$ and a standard deviation of $0.02$. We employ the Adam optimiser to train our network with the first and the second momentum setting to $0.5$ and $0.999$, respectively. 
We train the whole model for 100 epochs and the base learning rate is set to $2 \times 10^{-4}$ for the first 50 epochs and then we apply a linear decay strategy to decrease it to $0$ for the rest epochs. The batch size is set to $1$ for all experiments. The size of the dilated kernel $\tau$ in Eq.~(\ref{equ:dilation}) is experimentally set to 50 based on our ablation study on various values. For the data augmentation, we apply the random cropping and random flipping to avoid the over-fitting problem. The random cropping is implemented by first scaling each image to $448\times448$ and then randomly cropping a $400\times400$ region from the scaled image.
	
The network training involves all three sub-nets, and they impact each other through a forward or backward signal flow, e.g., the gradient from $\mathcal{R}$ can be propagated back to $\mathcal{I}$ and $\mathcal{G}$. While in the testing stage, given the shadow image and its shadow mask, only the shadow removal and refinement network are employed to produce the final shadow removal result with a resolution of $256 \times 256$ for evaluation. It approximately takes 16 hours to train the proposed G2R-ShadowNet on the ISTD dataset and 0.06 seconds to perform shadow removal for a test image.
    		
\begin{figure*}[htbp]\small
    \centering
	\begin{tabular}{ccccccccc}
		\hspace{-.2cm}
		\includegraphics[width=.122\textwidth]{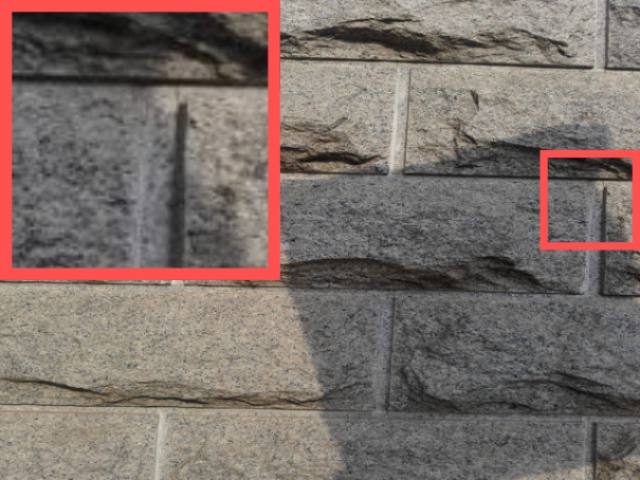} & \hspace{-.45cm}
		\includegraphics[width=.122\textwidth]{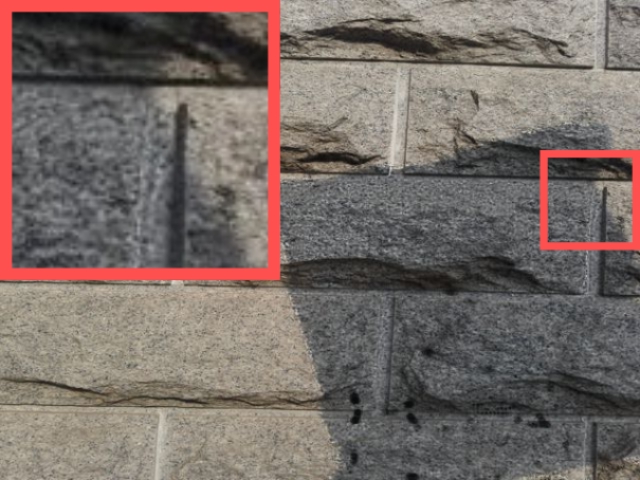} & \hspace{-.45cm}
		\includegraphics[width=.122\textwidth]{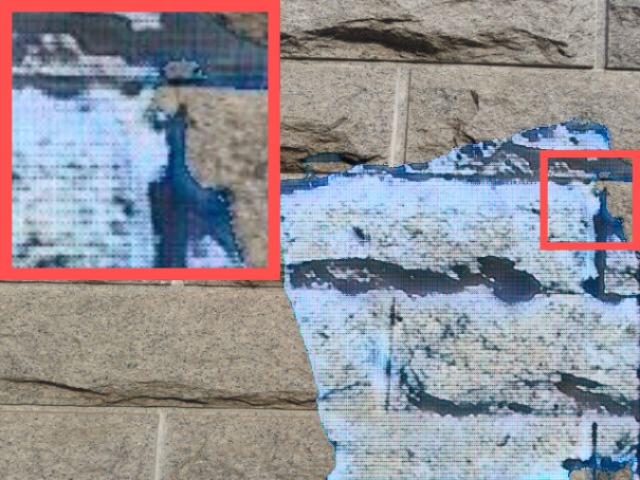} & \hspace{-.45cm}
		\includegraphics[width=.122\textwidth]{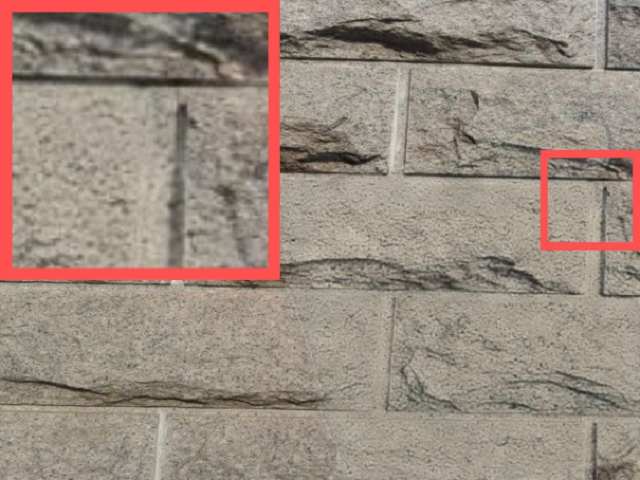} & \hspace{-.45cm}
		\includegraphics[width=.122\textwidth]{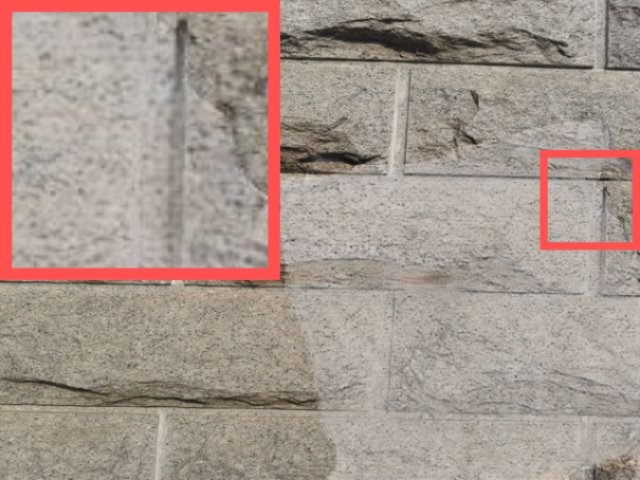} & \hspace{-.45cm}
		\includegraphics[width=.122\textwidth]{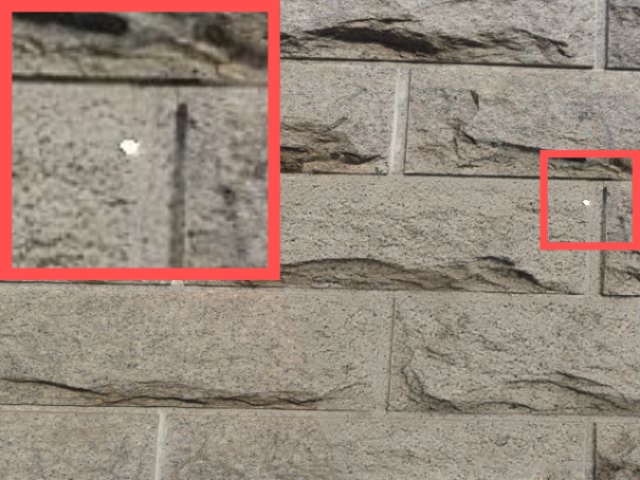} & \hspace{-.45cm}
		\includegraphics[width=.122\textwidth]{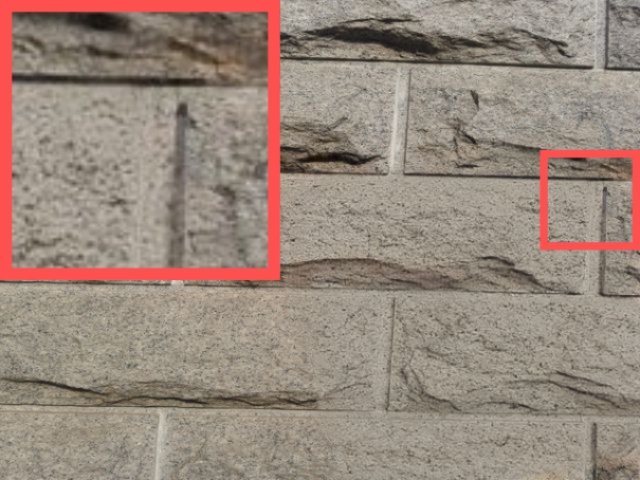} & \hspace{-.45cm}
		\includegraphics[width=.122\textwidth]{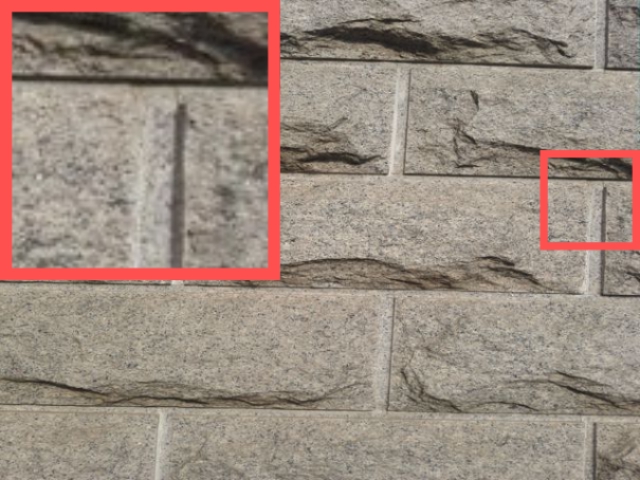}\vspace{-.06cm} \\
		\hspace{-.2cm} Input & \hspace{-.45cm}w/o $L_{GAN}$&\hspace{-.45cm}w/o $L_{iden}$ &\hspace{-.45cm} w/o $L_{rem}$&\hspace{-.45cm} w/o $L_{full}$&\hspace{-.45cm} w/o $L_{area}$&\hspace{-.45cm} Ours &\hspace{-.45cm} Ground truth \\
	\vspace{-7pt}
	\end{tabular}
    \caption{Visual comparisons for ablation study on the use of each loss term.}
    \label{fig:ablation}
	\vspace{-13pt}
\end{figure*}
    
\subsection{Ablation study}
To demonstrate the effectiveness of each key component of the proposed G2R-ShadowNet, we train and test several model variants on ISTD.
	
We first conduct an experiment to study each design of the refinement sub-net $\mathcal{R}$ by comparing it with the other two variants. One is obtained by removing $\mathcal{R}$ and the loss functions that are related to $\mathcal{R}$.
The other one is obtained without using the shadow mask $M$ as the input which means the region that is going to be refined is not known. The quantitative results are reported in Table~\ref{tab:a_r}.
The results indicate that the sub-net $\mathcal{R}$ plays quite an important role in our whole framework, which significantly improves the quality of the shadow removal result in terms of all three metrics. Including the shadow masks as the input is effective and brings improvements to all the metrics as well.
\begin{table}[htbp]\small
\centering
\vspace{-3pt}
\caption{Ablation study to verify the effectiveness of the refinement sub-net of our proposed G2R-ShadowNet on the test set of ISTD using all three evaluation metrics. Hereafter, `Shadow Region' represents that the metric is computed only on the shadow region of the image, and `All' represents that the metric is computed on the whole image. The best and the second best results are highlighted with \textbf{bold} font and \underline{underline}, respectively.}
\vspace{4pt}
\renewcommand\arraystretch{1.2}
\setlength{\tabcolsep}{1.3mm}{
	{\begin{tabular}{|l|ccc|ccc|}
			\hline
			\multirow{2}{*}{Method}&\multicolumn{3}{c|}{Shadow Region}& \multicolumn{3}{c|}{All}\\ 
			\cline{2-7} 
			&RMSE&PSNR&SSIM&RMSE&PSNR&SSIM\\ 
			\hline
			\hline
			Ours w/o $\mathcal{R}$ & \underline{12.3} & 29.20 & 0.975 & 4.6 & 26.04 & 0.913 \\
			\hline
			Ours w/o $M$ & 12.5 & \underline{29.68} & \underline{0.977} & \underline{4.3} & \underline{27.49} & \underline{0.940}\\
			\hline
			Ours & \textbf{8.9} & \textbf{33.58} &\textbf{0.979} &\textbf{3.9} &\textbf{30.52} & \textbf{0.944} \\
			\hline
\end{tabular}}}
\label{tab:a_r}
\vspace{-8pt}
\end{table}
	
Next, we perform an ablation study to verify the effectiveness of the joint training strategy of the three sub-nets in our method. Basically, the shadow generation, shadow removal, and the refinement of our proposed G2R-ShadowNet can impact the learning of each other through the backward signal flow. Therefore, we try to detach the result of each sub-net individually and train each variants one-by-one to see how one impacts the others. 
For instance, when the refinement result is detached, the back-propagated signal from the refinement sub-net $\mathcal{R}$ is not passed to $\mathcal{G}$ and $\mathcal{I}$.
	
The quantitative results are reported in Table~\ref{tab:ablation-detach}, from which the effectiveness of our designs are well justified. Particularly, we observe that the joint training of the three sub-nets (the last row) can boost the performance and achieve the highest score on both PSNR and SSIM compared with other variants. Especially, when results of all three sub-nets are detached, the performance drops in the shadow region on all the metrics.
By connecting only two of them for training, we can achieve certain performance gains. By comparing the second and third model variants, we find that the connection of $\mathcal{G}$ and $\mathcal{I}$ contributes more to the performance than the connection of $\mathcal{I}$ and $\mathcal{R}$.

\vspace{-4pt}
\begin{table}[htbp]\small
		\centering
		\caption{Ablation study to verify the effectiveness of the joint training strategy of the three sub-nets of our proposed G2R-ShadowNet on ISTD. `$\leftarrow$' and `$\nleftarrow$' in the first column denote the connection and detachment during back-propagation, respectively.}
	    \vspace{4pt}
	    \renewcommand\arraystretch{1.2}
	    \setlength{\tabcolsep}{1.3mm}{
			{\begin{tabular}{|l|ccc|ccc|}
					\hline
					\multirow{2}{*}{Method}&\multicolumn{3}{c|}{Shadow Region}& \multicolumn{3}{c|}{All}\\ 
					\cline{2-7} 
					&RMSE&PSNR&SSIM&RMSE&PSNR&SSIM\\ 
					\hline
					\hline
					Input image & 37.0 & 20.84 & 0.927 & 8.5 & 20.45 & 0.893\\
					\hline
					$\mathcal{G}\nleftarrow\mathcal{I}\nleftarrow\mathcal{R}$ & 9.3 & 33.43 & \underline{0.977} & \underline{3.9} & 30.24 & 0.941 \\
					\hline
					$\mathcal{G}\nleftarrow\mathcal{I}\leftarrow\mathcal{R}$ & 9.0 & 33.27 & \underline{0.977} & \textbf{3.8} & 30.36 & \underline{0.943} \\
					\hline
					$\mathcal{G}\leftarrow\mathcal{I}\nleftarrow\mathcal{R}$ & \textbf{8.8} & \underline{33.46} & \textbf{0.979} & \textbf{3.8} & \underline{30.48} & \textbf{0.944} \\
					\hline
					$\mathcal{G}\leftarrow\mathcal{I}\leftarrow\mathcal{R}$ & \underline{8.9} & \textbf{33.58} & \textbf{0.979} & \underline{3.9} & \textbf{30.52} & \textbf{0.944} \\
					\hline
		\end{tabular}}}
		\label{tab:ablation-detach}
	    \vspace{-8pt}
	\end{table}

We also conduct another ablation study to justify the effectiveness of each loss function by training our model without a specific loss term for each time. We report the quantitative results in Table~\ref{tab:ablation-loss}. From rows 1-2, we observe that the RMSE performance drops a lot without using $L_{GAN}$ and $L_{iden}$. The shadow removal loss $L_{rem}$ is also important and brings performance improvement in terms of all the metrics. When $L_{rem}$ is removed from the total loss, the shadow removal sub-net and the refinement sub-net are trained as a whole, which lacks individual constraints. Besides, removing $L_{full}$ leads to performance drop, which verifies the benefit of using the whole image as a constraint to train our G2R-ShadowNet. Finally, when $L_{area}$ is not calculated, the performance also slightly drops in the shadow region.
As shown in Fig.~\ref{fig:ablation}, the qualitative results are largely consistent with the above quantitative results in justifying the effectiveness of each loss term. Compared with the model training by combining all the loss terms, the other variants that are trained based on a subset of loss terms may cause obvious artefacts on the results, e.g., a white area in the shadow edge as shown in Fig.~\ref{fig:ablation} (column 6).
\begin{table}[htbp]\small
	\centering
    \vspace{-4pt}
	\caption{Ablation study on the choices of the loss functions for the proposed G2R-ShadowNet.}
    \vspace{4pt}
    \renewcommand\arraystretch{1.2}
    \setlength{\tabcolsep}{1.0mm}{
		{\begin{tabular}{|l|ccc|ccc|}
				\hline
				\multirow{2}{*}{Method}&\multicolumn{3}{c|}{Shadow Region}&\multicolumn{3}{c|}{All}\\
				\cline{2-7} 
				 & RMSE & PSNR & SSIM & RMSE & PSNR & SSIM \\ 
				\hline
				\hline
				Ours w/o $L_{GAN}$ & 42.1 & 19.62 & 0.911 & 9.3 & 19.27 & 0.878 \\\hline
				Ours w/o $L_{iden}$ & 43.1 & 21.57 & 0.878 & 9.8 & 20.70 & 0.825\\\hline
				Ours w/o $L_{rem}$ & 9.8 & 32.71 & \underline{0.977} & 4.1 & 29.91 & 0.942 \\\hline
				Ours w/o $L_{full}$ & 12.3 & 29.78 & 0.966 & 4.3 & 27.66 & 0.923 \\\hline
				Ours w/o $L_{area}$ & \underline{9.3} & \underline{33.22} & \textbf{0.979} & \textbf{3.8} & \underline{30.40} & \underline{0.943} \\\hline
				Ours & \textbf{8.9} & \textbf{33.58} & \textbf{0.979} & \underline{3.9} & \textbf{30.52} & \textbf{0.944} \\\hline
	\end{tabular}}}
	\label{tab:ablation-loss}
    \vspace{-8pt}
\end{table}
	
We also carry out a set of experiments to explore the impact of choosing different dilated kernel size $\tau$ in $L_{area}$. Specifically, we set $\tau$ to $0$, $5$, $15$, $50$, and $100$ and train our model, respectively. Quantitative results are reported in Table~\ref{tab:ablation-dilation}, which shows that 50 is the optimal dilated kernel size that help achieve the best performance.
\vspace{-4pt}
\begin{table}[htbp]\small
		\centering
		\caption{Influence of the dilation kernel $\tau$ in $L_{area}$ to the performance of the proposed G2R-ShadowNet.}
	    \vspace{4pt}
	    \renewcommand\arraystretch{1.2}
	    \setlength{\tabcolsep}{1.7mm}{
			{\begin{tabular}{|l|ccc|ccc|}
				\hline
				\multirow{2}{*}{Method}&\multicolumn{3}{c|}{Shadow Region}&\multicolumn{3}{c|}{All}\\
				\cline{2-7} 
				 & RMSE & PSNR & SSIM & RMSE & PSNR & SSIM \\ 
				\hline
				\hline
				$\tau=0$ & 9.5 & 33.01 & 0.977 & \textbf{3.9} & 30.11 & 0.939 \\
				\hline
				$\tau=5$ & 9.5 & 32.70 & 0.976 & \underline{4.0} & 29.89 & 0.939 \\
				\hline
				$\tau=15$ & \underline{9.1} & \underline{33.54} & \textbf{0.980} & \textbf{3.9} & \underline{30.40} & \textbf{0.944} \\
				\hline
				$\tau=50$ & \textbf{8.9} & \textbf{33.58} & \underline{0.979} & \textbf{3.9} & \textbf{30.52} & \textbf{0.944} \\
				\hline
				$\tau=100$ & 9.4 & 33.08 & 0.978 & \textbf{3.9} & 30.16 & \underline{0.943} \\
				\hline
		\end{tabular}}}
		\label{tab:ablation-dilation}
	    \vspace{-8pt}
	\end{table}
\begin{table*}[htbp]\small
		\centering
		\caption{Quantitative comparison results of the proposed G2R-ShadowNet with the state-of-the-art methods. `Non-Shadow Region’ indicates that RMSE is computed on the non-shadow region of the testing images. `RMSE*' indicates that RMSE is calculated by averaging the RMSE of all pixels in the shadow regions over the whole testing set. The results of these methods are either obtained from their original publications or produced by us using their official codes (marked with `$^\star$').
		}
	    \vspace{4pt}
	    \renewcommand\arraystretch{1.2}
	    \setlength{\tabcolsep}{1.45mm}{
			{\begin{tabular}{|l|l|cccc|ccc|ccc|}
					\hline
					\multirow{2}{*}{Method}&\multirow{2}{*}{Training Data}&\multicolumn{4}{c|}{Shadow Region}&\multicolumn{3}{c|}{Non-Shadow Region}& \multicolumn{3}{c|}{All}\\ 
					\cline{3-12} 
					&&RMSE*&RMSE&PSNR&SSIM&RMSE&PSNR&SSIM&RMSE&PSNR&SSIM\\ 
					\hline
					\hline
					Yang \etal~\cite{yang2012shadowremoval} & - &24.7& 23.2 & 21.57 & 0.878  & 14.2 & 22.25 & 0.782 & 15.9 & 20.26 & 0.706  \\
					Gong and Cosker~\cite{gong2014interactive} & - &13.3& 13.0  & 30.53 & 0.972  & \textbf{2.6} & \textbf{36.63} & \textbf{0.982} & 4.3 & 28.96 & 0.943 \\
					\hline
					Guo \etal~\cite{guo2012paired} & Shd.Free+Shd.Mask &22.0& 20.1 & 26.89 & 0.960  & 3.1 & 35.48 & 0.975 & 6.1 & 25.51 & 0.924 \\
					ST-CGAN~\cite{wang2018stacked} & Shd.Free+Shd.Mask &13.4& 12.0 & 31.70 & 0.979  & 7.9 & 26.39 & 0.956 & 8.6 & 24.75 & 0.927 \\
					SP+M-Net~\cite{Le2019Shadow} & Shd.Free+Shd.Mask & \underline{7.9} & \underline{8.1} & \underline{35.08} & \underline{0.984} & \underline{2.8} & \underline{36.38} & \underline{0.979} & \textbf{3.6} & \underline{31.89} & \underline{0.953} \\
        			G2R-ShadowNet \textit{Sup.} & Shd.Free+Shd.Mask & \textbf{7.3} & \textbf{7.9} & \textbf{36.12} & \textbf{0.988} & 2.9 & 35.21 & 0.977 & \textbf{3.6} & \textbf{31.93} & \textbf{0.957} \\
					\hline
					Mask-ShadowGAN$^\star$~\cite{wang2018stacked} & Shd.Free (Unpaired) &9.9& 10.8 & 32.19 & \underline{0.984}  & 3.8 & 33.44 & 0.974 & 4.8 & 28.81 & 0.946 \\
					LG-ShadowNet~\cite{liu2021shadow} & Shd.Free (Unpaired) &9.7& 9.9 & 32.44 & 0.982  & 3.4 & 33.68 & 0.971 & 4.4 & 29.20 & 0.945 \\
					\hline
					Le and Samaras~\cite{le2020from} & Shd.Mask & 9.7 & 10.4 & 33.09 & 0.983 & 2.9 & 35.26 & 0.977 & 4.0 & 30.12 & 0.950 \\
					\bf{G2R-ShadowNet} & Shd.Mask & 8.8 & 8.9 & 33.58 & 0.979 & 2.9 & 35.52 & 0.976 & \underline{3.9} & 30.52 & 0.944 \\
					\hline
		\end{tabular}}}
		\label{tab:sota}
	    \vspace{-8pt}
	\end{table*}

\begin{figure}[htbp]\small
    \centering
	\begin{tabular}{cccc}
		\hspace{-.2cm}\includegraphics[width=.117\textwidth]{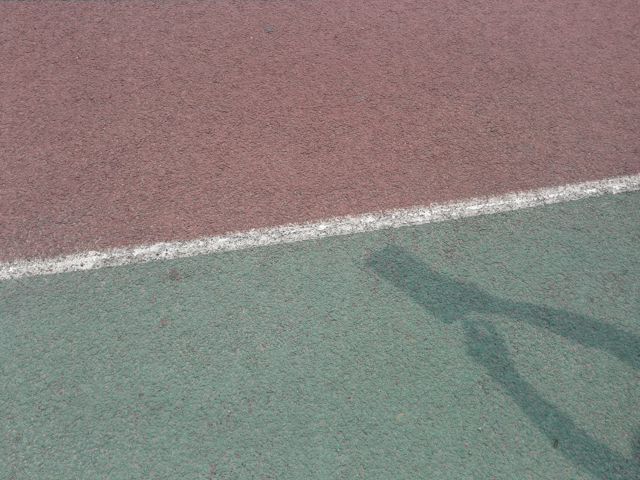} & \hspace{-.45cm}
		\includegraphics[width=.117\textwidth]{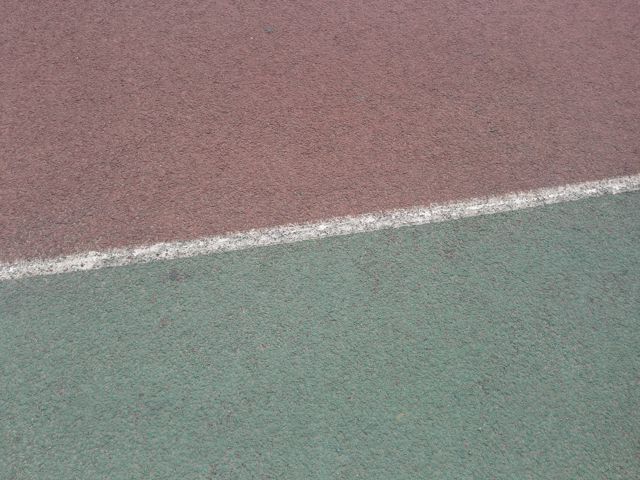}& \hspace{-.45cm}
		\includegraphics[width=.117\textwidth]{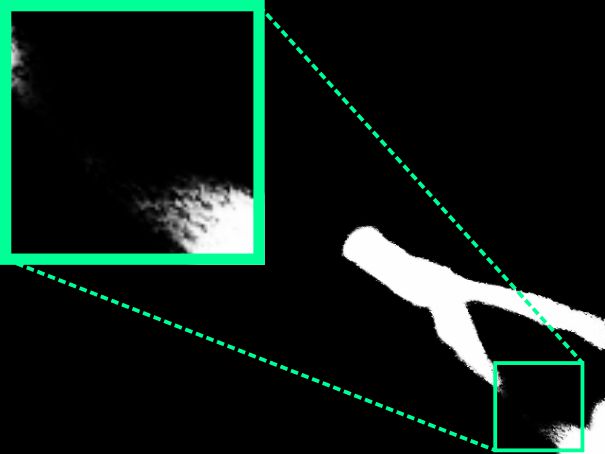}& \hspace{-.45cm}
		\includegraphics[width=.117\textwidth]{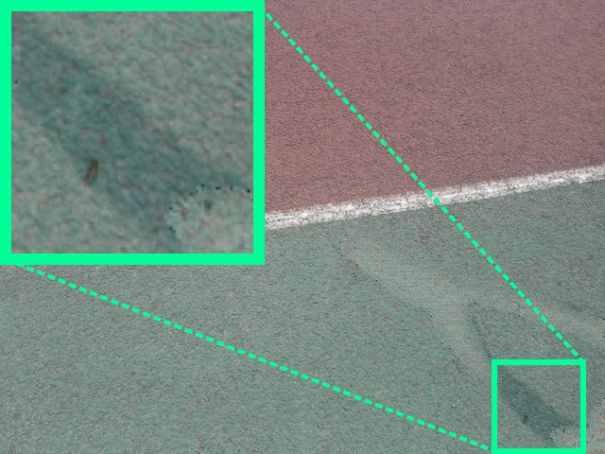}\vspace{-.06cm} \\
		\hspace{-.2cm}\includegraphics[width=.117\textwidth]{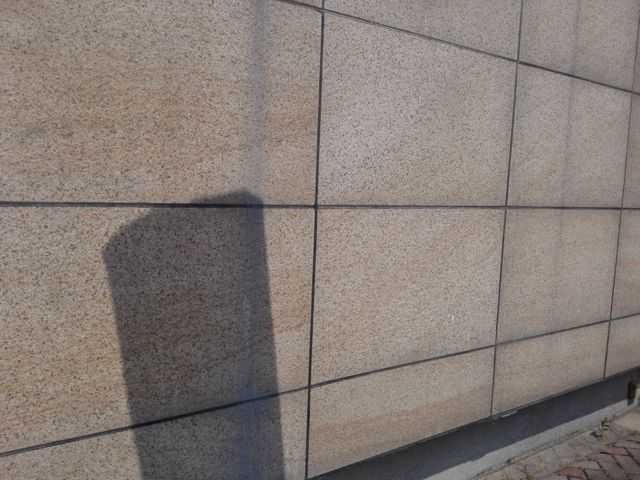} & \hspace{-.45cm}
		\includegraphics[width=.117\textwidth]{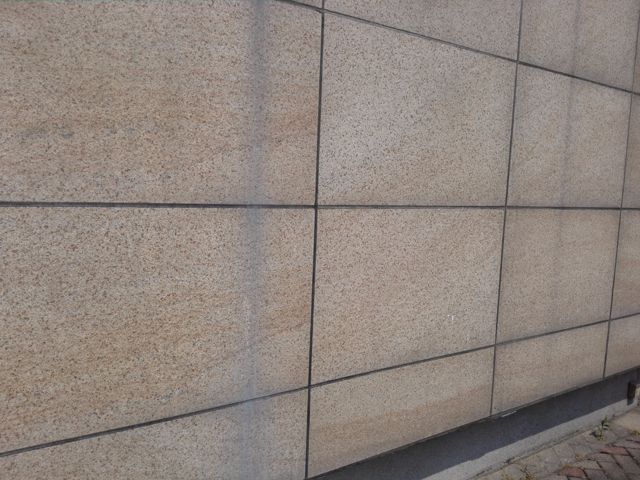}& \hspace{-.45cm}
		\includegraphics[width=.117\textwidth]{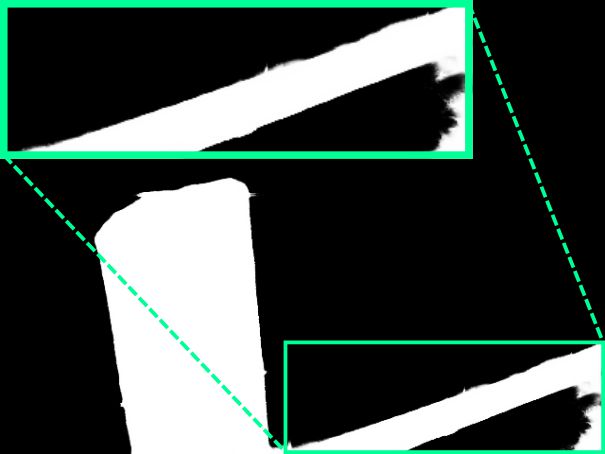}& \hspace{-.45cm}
		\includegraphics[width=.117\textwidth]{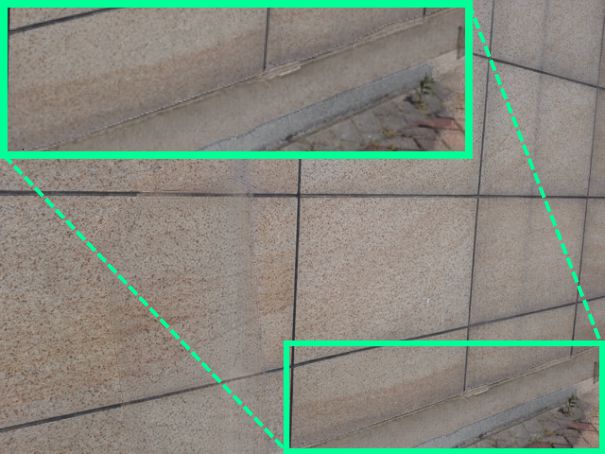}\vspace{-.09cm}\\
		\hspace{-.2cm} Input & \hspace{-.45cm} \scriptsize Ground truth &\hspace{-.45cm} \scriptsize Detected mask & \hspace{-.45cm} Ours \\  
	\end{tabular}
    \caption{Failure cases on the ISTD dataset. The detected masks are generated by the shadow detector~\cite{zhu2018bidirectional}.}
    \label{fig:fail}
	\vspace{-8pt}
\end{figure}

It is worth to mention that the performance of shadow removal is highly affected by the predicted shadow mask obtained via~\cite{zhu2018bidirectional}. We show some failure cases caused by the false detected shadow masks in Fig.~\ref{fig:fail}. If a shadow is not detected, it cannot be removed, as shown in the top of Fig.~\ref{fig:fail}. If a non-shadow region is mis-detected as shadow, it may become brighter after shadow removal, as shown in the bottom of Fig.~\ref{fig:fail}.
If we use the ground-truth shadow masks as input for testing, the performance of our model can be further improved. We conduct this experiment and find that RMSE, PSNR and SSIM of the predicted results from our proposed method can reach 8.6, 34.01, and 0.979, respectively, when evaluated on the shadow region of ISTD.
	
\subsection{Comparison with the state-of-the-arts}
In this subsection, we compare our proposed weakly-supervised method with several state-of-the-art methods on ISTD. In addition, we train our method on paired data by skipping the generation and training directly with shadow/non-shadow images from ISTD, and denote this model as G2R-ShadowNet \textit{Sup.}.

The methods that we are compared with include Gong and Cosker~\cite{gong2014interactive}, Guo \etal~\cite{guo2012paired}, Yang \etal~\cite{yang2012shadowremoval}, ST-CGAN~\cite{wang2018stacked}, Mask-ShadowGAN~\cite{hu2019mask}, SP+M-Net~\cite{Le2019Shadow}, LG-ShadowNet~\cite{liu2021shadow}, and Le and Samaras~\cite{le2020from} on ISTD. Among them, Guo \etal~\cite{guo2012paired}, Yang \etal~\cite{yang2012shadowremoval}, and Gong and Cosker~\cite{gong2014interactive} use the pre-calculated image priors for shadow removal.
ST-CGAN~\cite{wang2018stacked} and SP+M-Net~\cite{Le2019Shadow} leverage paired shadow and shadow-free images, as well as shadow masks to train their models. Mask-ShadowGAN~\cite{hu2019mask} and LG-ShadowNet~\cite{liu2021shadow} require unpaired shadow and shadow-free images for training. Le and Samaras~\cite{le2020from} needs shadow images and shadow masks to train their network and our method use the same type of data as Le and Samaras~\cite{le2020from}.
	\begin{figure*}[htbp]\small
	    \centering
		\begin{tabular}{ccccccc}
			\hspace{-.2cm}
			\includegraphics[width=.139\textwidth]{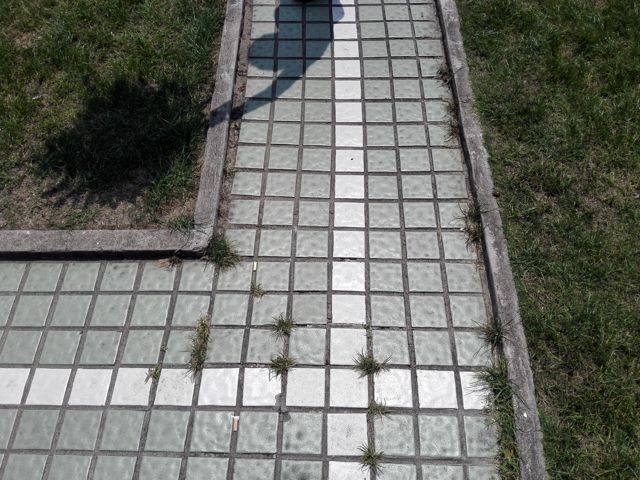} & \hspace{-.45cm}
			\includegraphics[width=.139\textwidth]{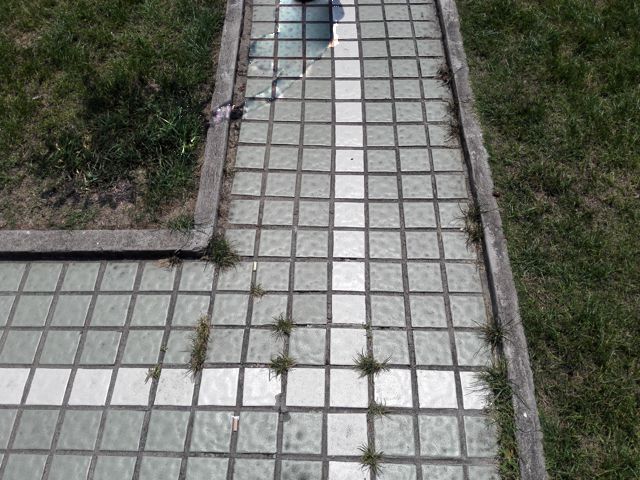} & \hspace{-.45cm}
			\includegraphics[width=.139\textwidth]{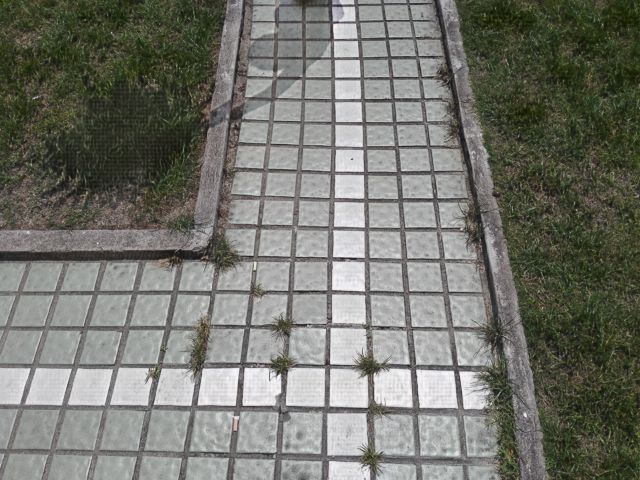} & \hspace{-.45cm}
			\includegraphics[width=.139\textwidth]{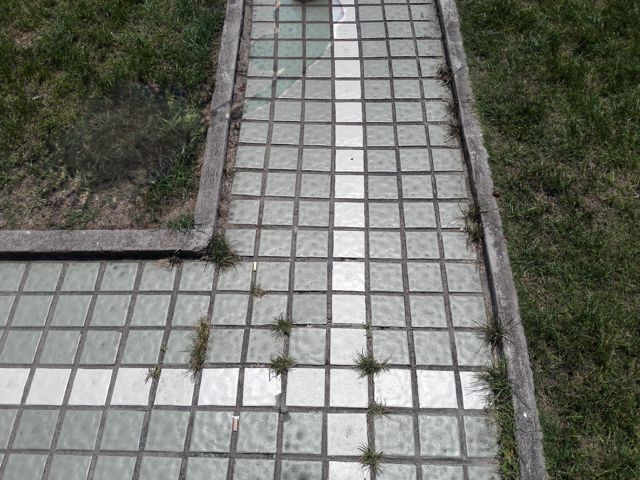} & \hspace{-.45cm}
			\includegraphics[width=.139\textwidth]{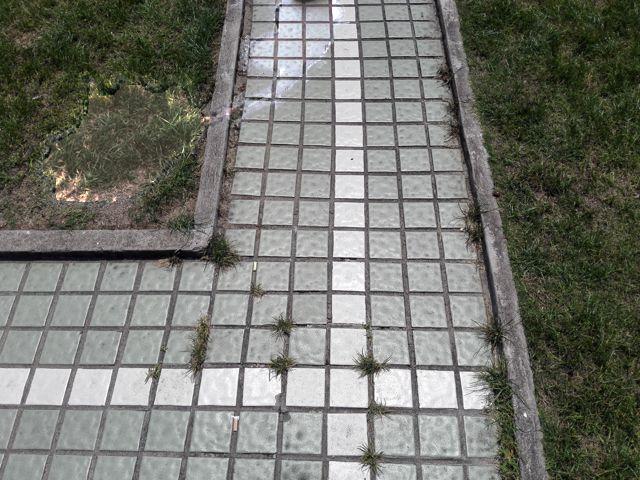} & \hspace{-.45cm}
			\includegraphics[width=.139\textwidth]{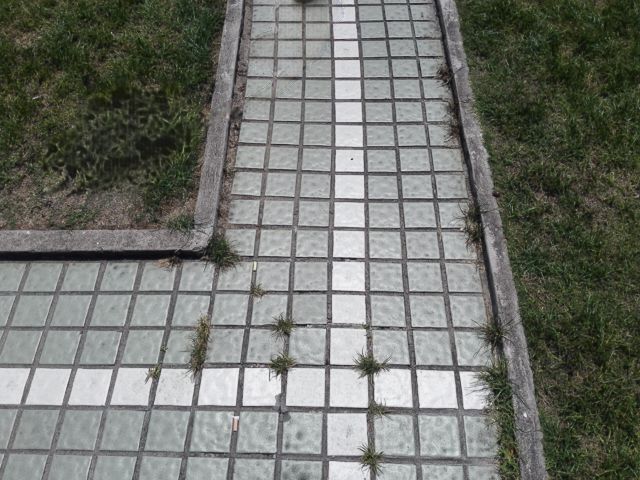} & \hspace{-.45cm}
			\includegraphics[width=.139\textwidth]{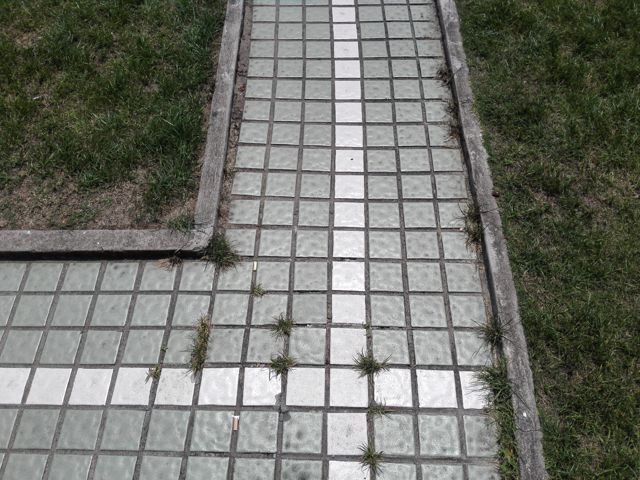}\vspace{-.06cm} \\
			\hspace{-.2cm}
			\includegraphics[width=.139\textwidth]{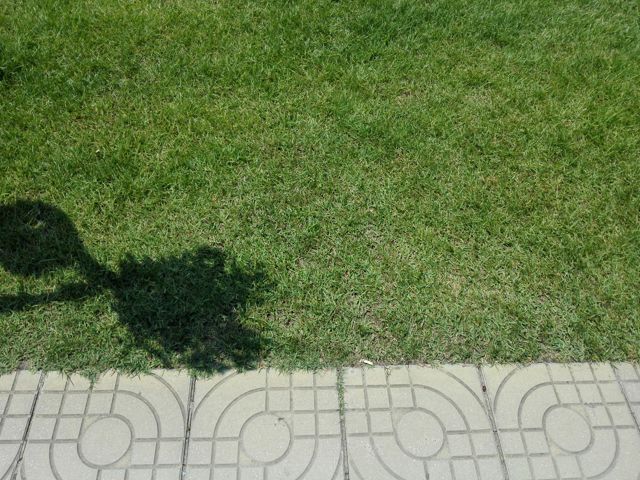} & \hspace{-.45cm}
			\includegraphics[width=.139\textwidth]{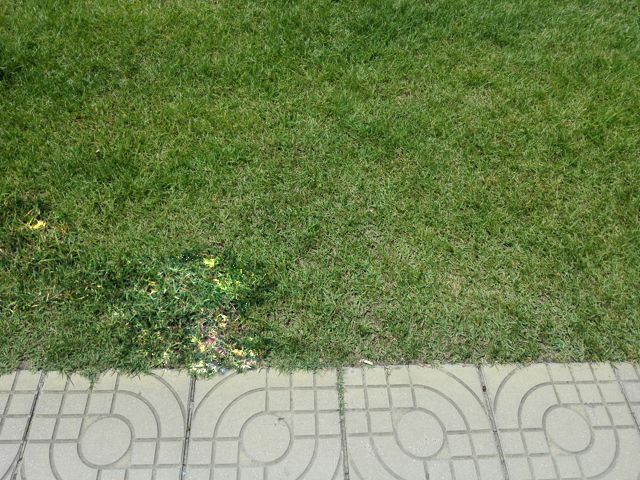} & \hspace{-.45cm}
			\includegraphics[width=.139\textwidth]{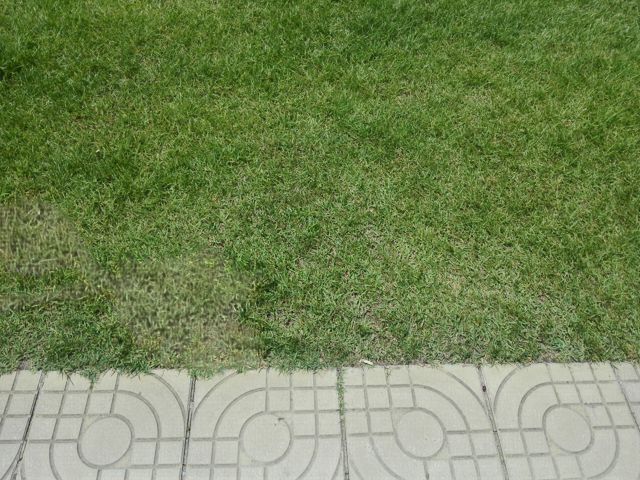} & \hspace{-.45cm}
			\includegraphics[width=.139\textwidth]{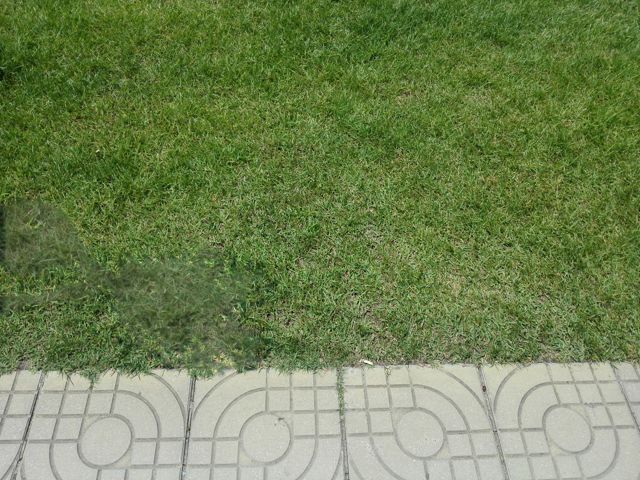} & \hspace{-.45cm}
			\includegraphics[width=.139\textwidth]{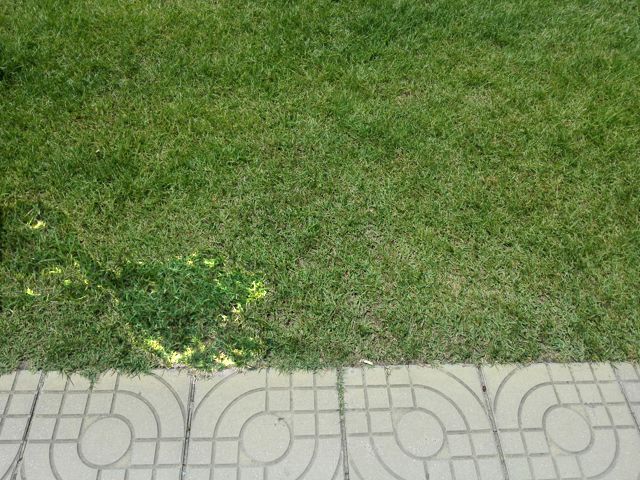} & \hspace{-.45cm}
			\includegraphics[width=.139\textwidth]{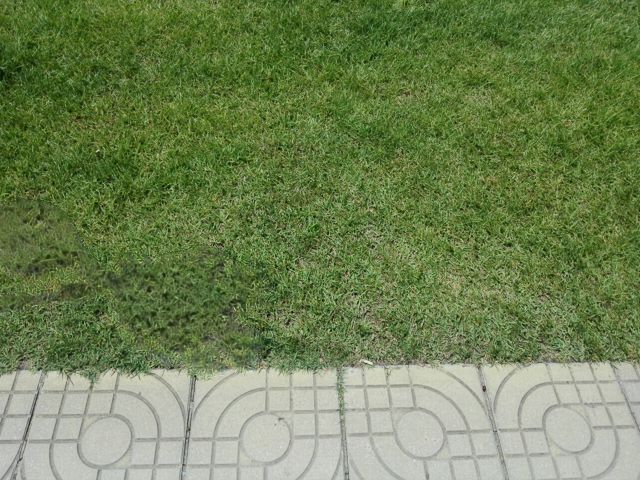} & \hspace{-.45cm}
			\includegraphics[width=.139\textwidth]{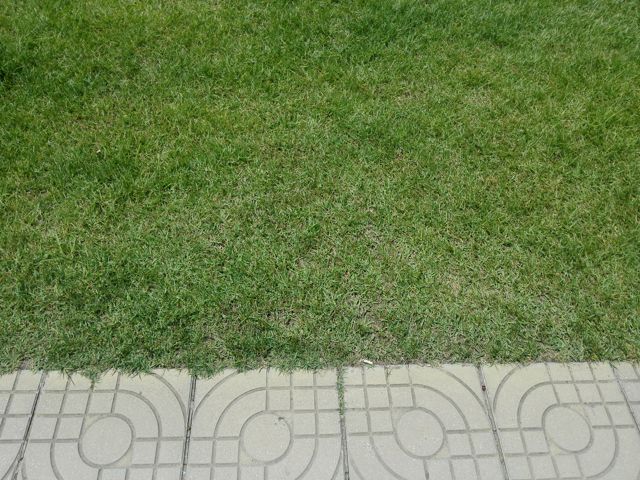}\vspace{-.06cm} \\
			\hspace{-.2cm}
			\includegraphics[width=.139\textwidth]{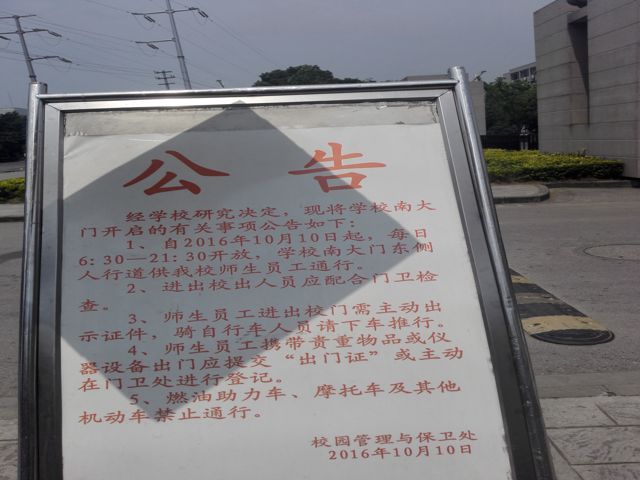} & \hspace{-.45cm}
			\includegraphics[width=.139\textwidth]{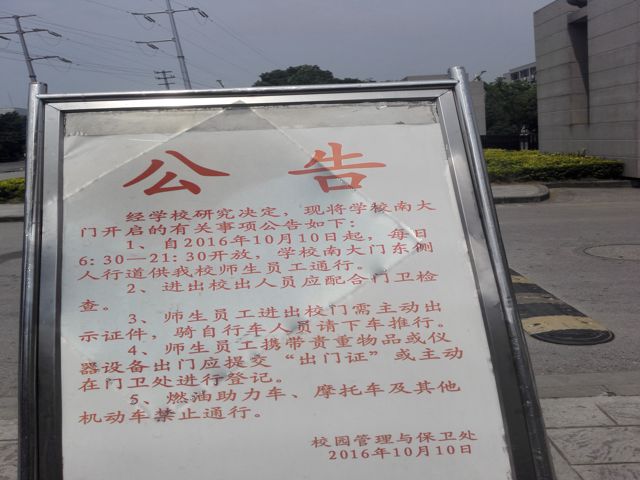} & \hspace{-.45cm}
			\includegraphics[width=.139\textwidth]{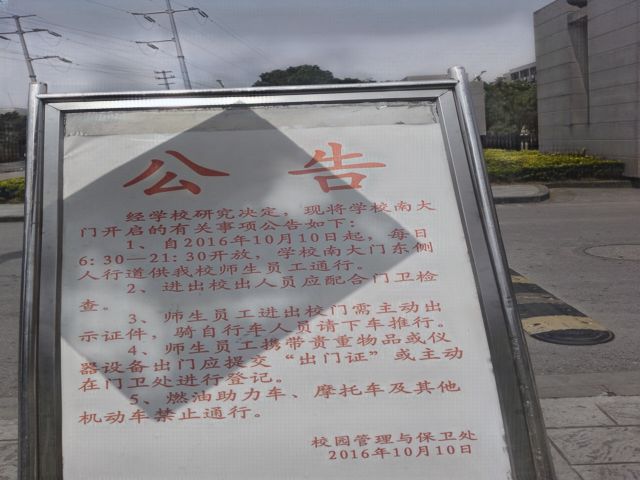} & \hspace{-.45cm}
			\includegraphics[width=.139\textwidth]{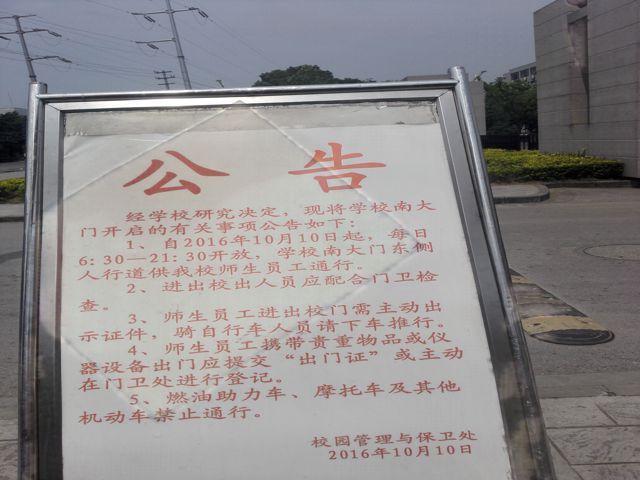} & \hspace{-.45cm}
			\includegraphics[width=.139\textwidth]{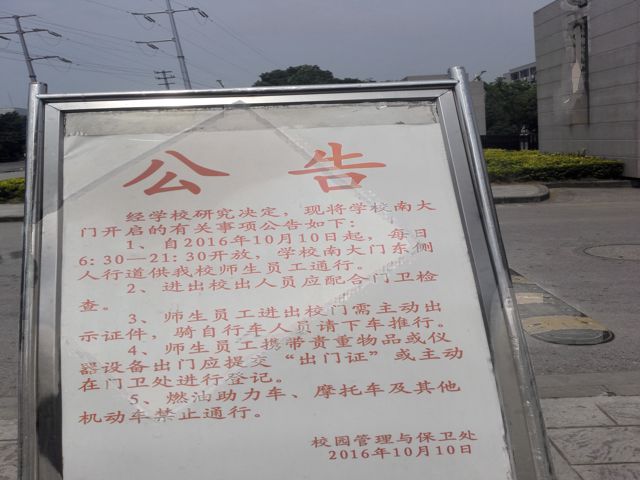} & \hspace{-.45cm}
			\includegraphics[width=.139\textwidth]{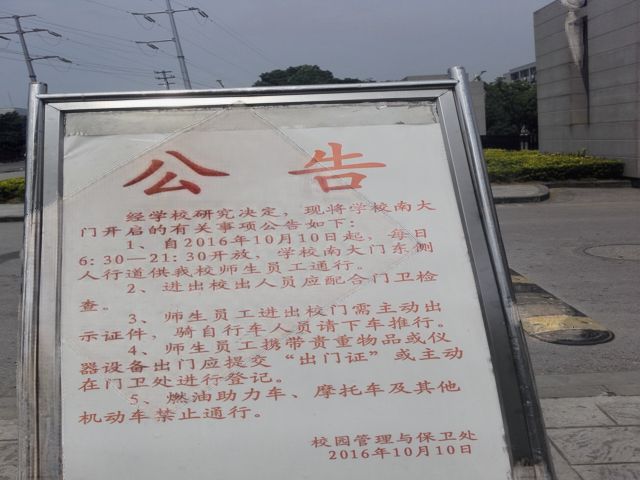} & \hspace{-.45cm}
			\includegraphics[width=.139\textwidth]{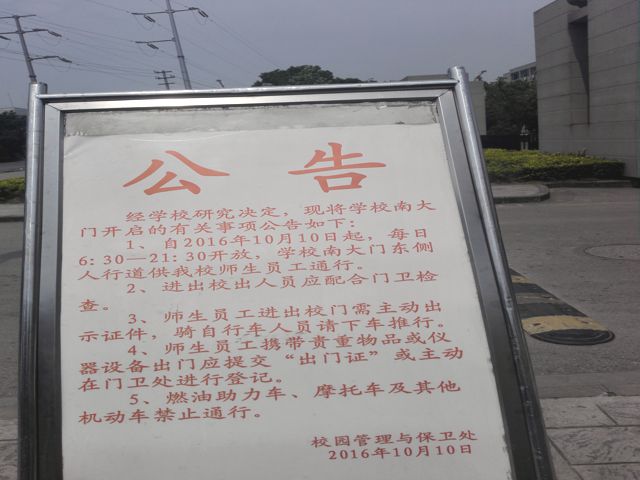}\vspace{-.06cm} \\
			\hspace{-.2cm}
			\includegraphics[width=.139\textwidth]{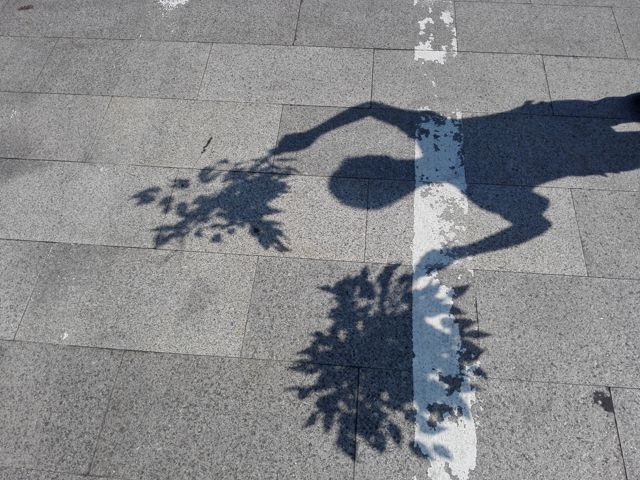} & \hspace{-.45cm}
			\includegraphics[width=.139\textwidth]{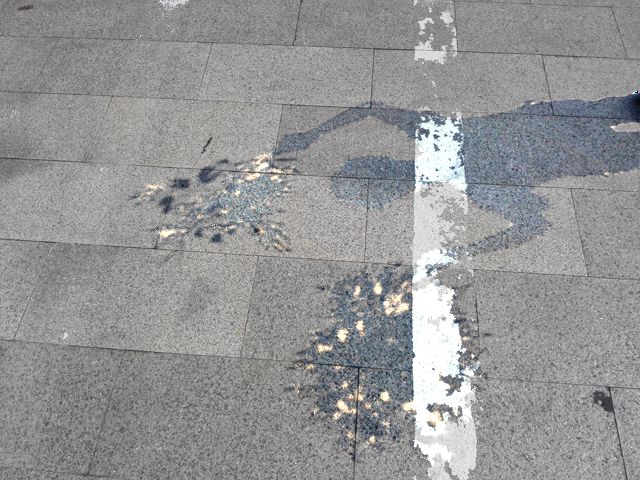} & \hspace{-.45cm}
			\includegraphics[width=.139\textwidth]{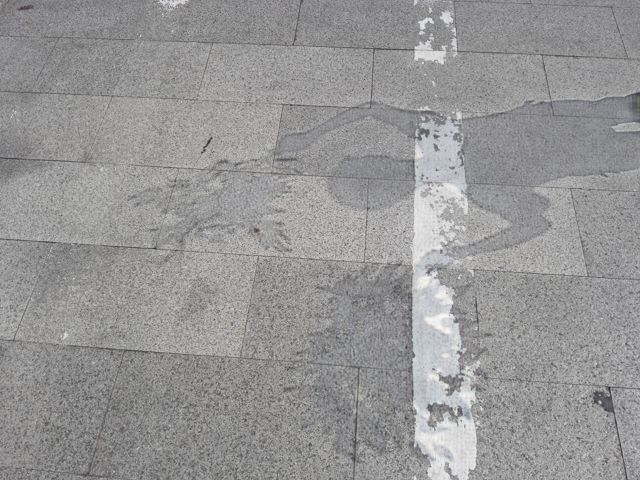} & \hspace{-.45cm}
			\includegraphics[width=.139\textwidth]{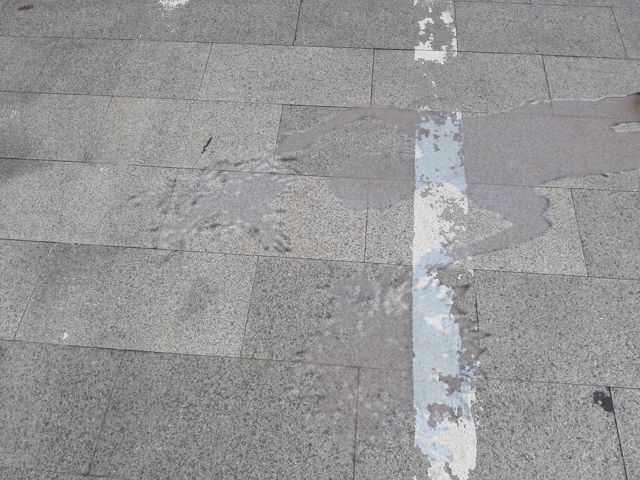} & \hspace{-.45cm}
			\includegraphics[width=.139\textwidth]{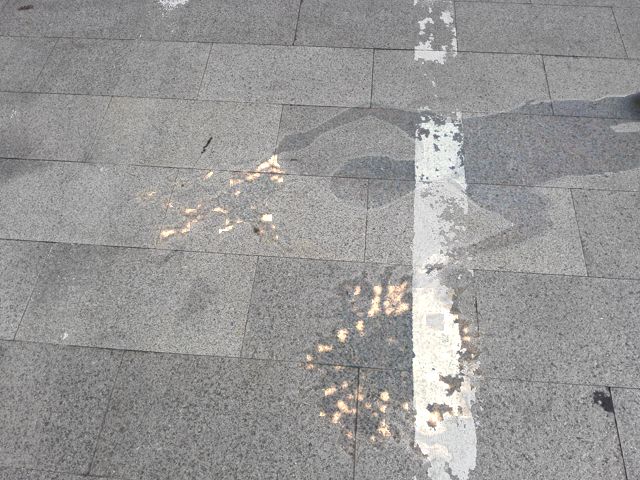} & \hspace{-.45cm}
			\includegraphics[width=.139\textwidth]{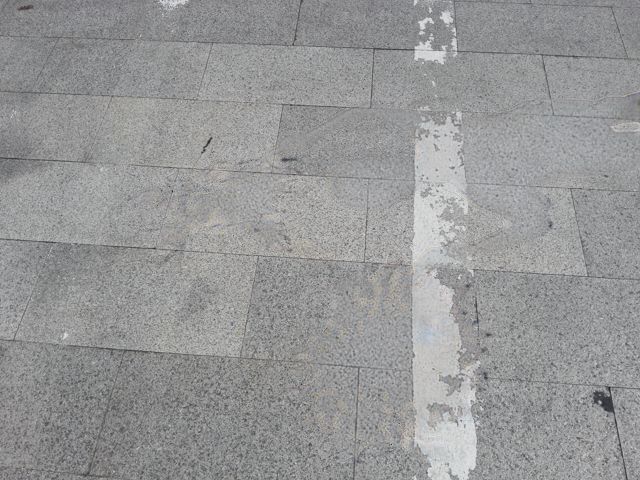} & \hspace{-.45cm}
			\includegraphics[width=.139\textwidth]{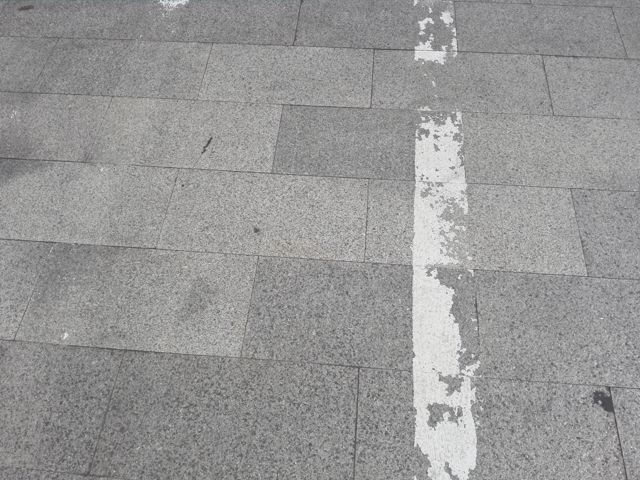}\vspace{-.06cm} \\
			\hspace{-.2cm} Input frame & \hspace{-.45cm} {\footnotesize Gong and Cosker}~\cite{gong2014interactive}&\hspace{-.45cm}
			{\footnotesize LG-ShadowNet~\cite{liu2021shadow}} &\hspace{-.45cm} SP+M-Net~\cite{Le2019Shadow} &\hspace{-.45cm} {\footnotesize Le and Samaras}~\cite{le2020from} &\hspace{-.45cm}
			Ours &\hspace{-.45cm} Ground truth \\
			
		\vspace{-5pt}
		\end{tabular}
        \caption{Visualisation comparisons on four challenging samples from the testing set of the ISTD dataset.}
        \label{fig:sota}
	    \vspace{-8pt}
    \end{figure*}

Quantitative results are shown in Table~\ref{tab:sota}. From the first block, we observe that our method outperforms the methods using the pre-calculated image priors except for the non-shadow region.
Note that Gong and Cosker~\cite{gong2014interactive} use an interactive method, which requires user input, to define the shadow and non-shadow regions in testing, while we only use the one automatically generated by \cite{zhu2018bidirectional}. 
The methods in the second block share the same type of training data, including both shadow-free images and the shadow masks. Our weakly-supervised method achieves competitive performance with these methods but using less training data: we do not use shadow-free images.
We also observe that our method trained on paired data, i.e., G2R-ShadowNet \textit{Sup.}, outperforms all the methods in the shadow region and the whole image.
In the third block, both Mask-ShadowGAN and LG-ShadowNet train their shadow removal models using unpaired shadow and shadow-free images. We can see that our method outperforms these two methods.

The comparison to Le and Samaras~\cite{le2020from} is more fair since it is the only previous work that trains the model without using shadow-free images, which is also main goal of our method. From the last block, we can see that our method outperforms~\cite{le2020from} on most metrics except that two of the SSIM values are slightly below ~\cite{le2020from}.
    
Figure~\ref{fig:sota} shows the qualitative results of our method and the other state-of-the-art methods on four challenging samples drawn from the testing set of ISTD. Compared with other methods, our method can produce results with less artefacts. Moreover, the colour in the shadow region is more consistent with the surrounding area using our method, while the patch-based method~\cite{le2020from} tends to produce over-lightened colour in the non-shadow region (column 5), making them easy to distinguish even after the shadow removal.

\subsection{Generalisation ability}
Finally, we show the generalisation ability of the proposed G2R-ShadowNet by comparing it with Mask-ShadowGAN~\cite{hu2019mask}, SP+M-Net~\cite{Le2019Shadow}, LG-ShadowNet~\cite{liu2021shadow}, and Le and Samaras~\cite{le2020from}. Here all methods are trained on ISTD and tested on the video dataset without additional training or fine-tuning. The quantitative results are reported in Table~\ref{tab:video}.

\begin{table}[htbp]\small
	\centering
	\caption{Quantitative comparison of the generalisation ability of the proposed G2R-ShadowNet and the state-of-the-art methods on the video shadow removal dataset. Note that we compute the metrics only in the moving-shadow region. `RMSE$\dagger$' is the RMSE computed by using the moving-shadow mask with a threshold of 40, while other metrics are computed using a threshold of 80. `-' in the first two rows means the results are not publicly available.
	} 
    \vspace{5pt}
    \renewcommand\arraystretch{1.2}
    \setlength{\tabcolsep}{1.8mm}{
	\begin{tabular}{|l|cccc|}
			\hline
			Method & RMSE & RMSE$\dagger$ & PSNR & SSIM \\ 
			\hline
			\hline
			SP+M-Net~\cite{Le2019Shadow} & -  & 22.2 & - & - \\
			\hline
			Le and Samaras~\cite{le2020from} & - & 20.9 & - & - \\
			\hline
			Mask-ShadowGAN$^\star$~\cite{hu2019mask} & 22.7 & 19.6 & 20.38 & \textbf{0.887} \\
			\hline
			LG-ShadowNet$^\star$~\cite{liu2021shadow} & \underline{22.0} & \textbf{18.3} & \underline{20.68} & 0.880 \\
			\hline
			\textbf{G2R-ShadowNet (Ours)} & \textbf{21.8}  & \underline{18.8} & \textbf{21.07} & \underline{0.882} \\
			\hline
	\end{tabular}}
	\label{tab:video}
    \vspace{-8pt}
\end{table}

We observe that our method outperforms Mask-ShadowGAN~\cite{hu2019mask} and LG-ShadowNet~\cite{liu2021shadow} significantly on RMSE and PSNR metrics, indicating that our method has better generalisation ability on other unseen environments. We also fine-tune our model on each testing video for additional 1 epoch and it further improves the performance gains on RMSE by about 14\% (from 21.8 to 18.7).
	
\begin{figure}[htbp]\small
    \centering
	\begin{tabular}{cccc}
		\hspace{-.2cm}\includegraphics[width=.117\textwidth]{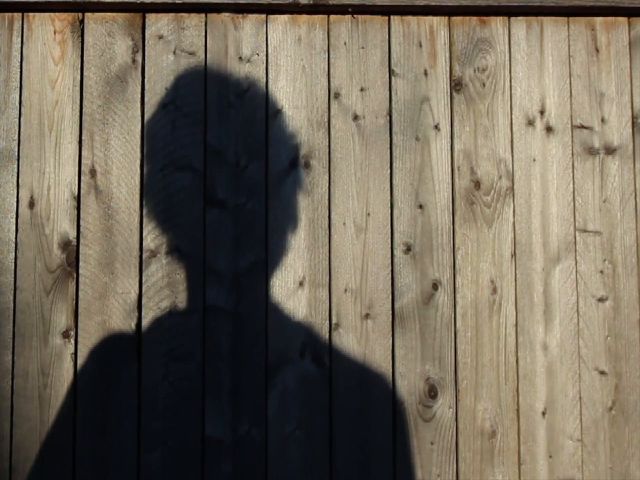} & \hspace{-.45cm}
		\includegraphics[width=.117\textwidth]{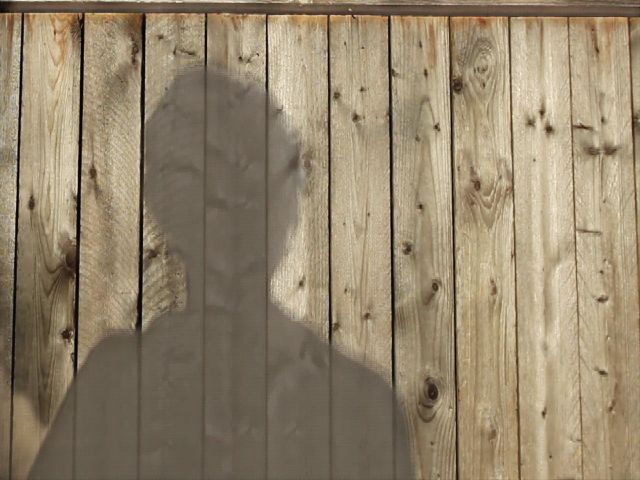}& \hspace{-.45cm}
		\includegraphics[width=.117\textwidth]{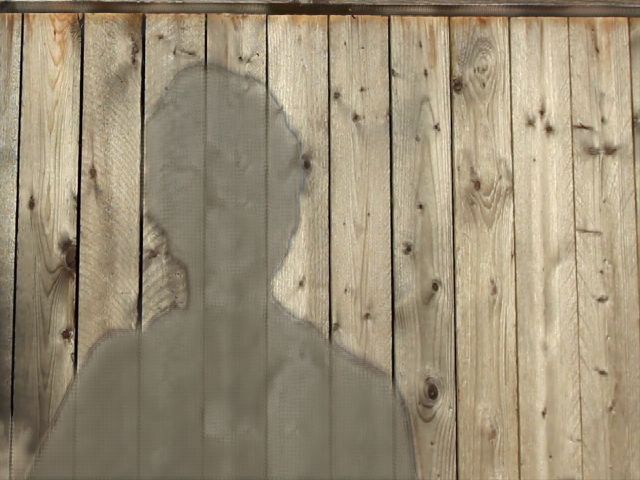}& \hspace{-.45cm}
		\includegraphics[width=.117\textwidth]{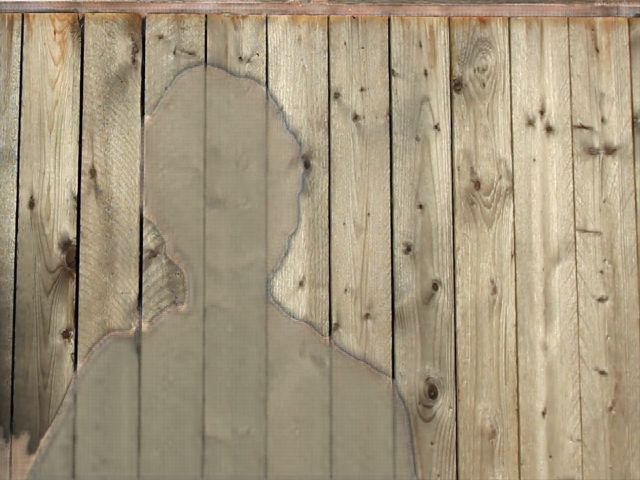}\vspace{-.06cm} \\
		\hspace{-.2cm}\includegraphics[width=.117\textwidth]{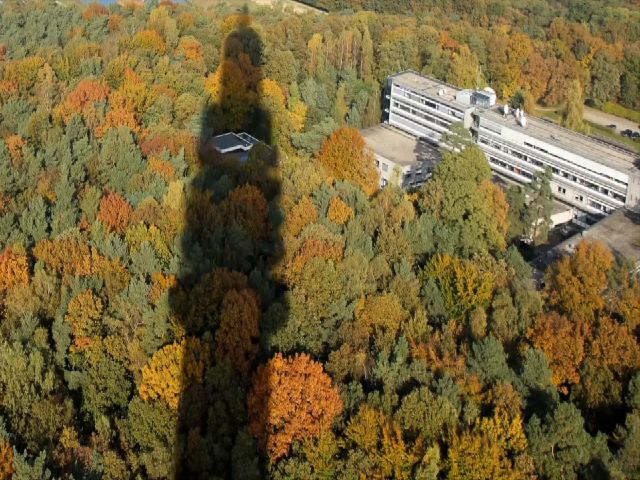} & \hspace{-.45cm}
		\includegraphics[width=.117\textwidth]{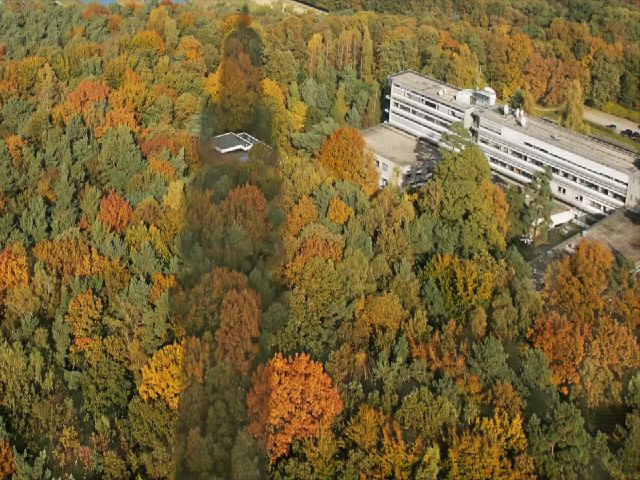}& \hspace{-.45cm}
		\includegraphics[width=.117\textwidth]{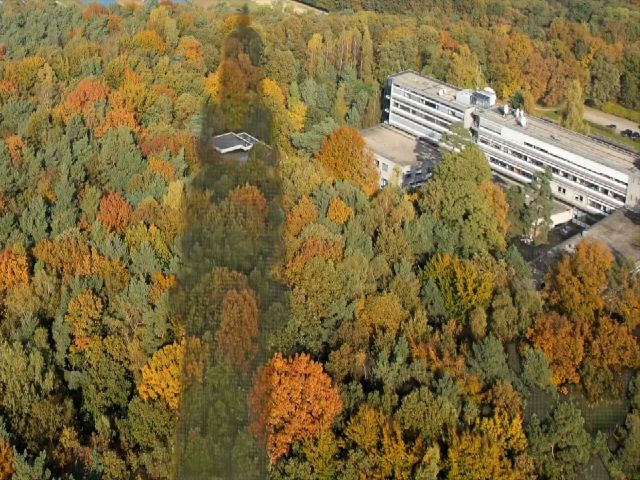}& \hspace{-.45cm}
		\includegraphics[width=.117\textwidth]{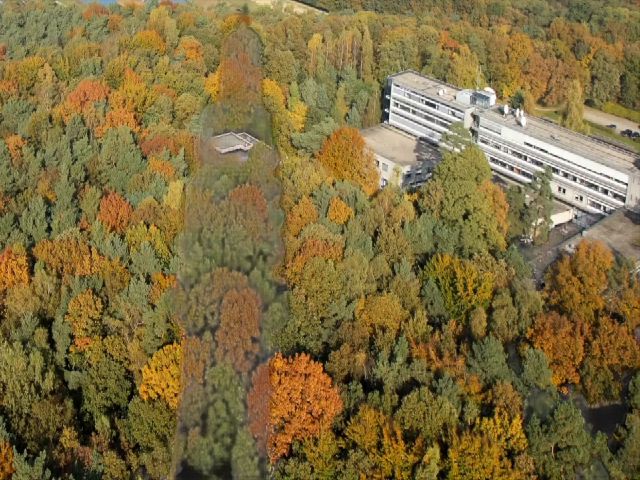}\vspace{-.06cm}\\
		\hspace{-.2cm} Input frame & \hspace{-.45cm} {\scriptsize Mask-ShadowGAN} &\hspace{-.45cm} {\footnotesize LG-ShadowNet} & \hspace{-.45cm} Ours \\  
	\end{tabular}
	\vspace{1pt}
    \caption{Visualisation comparisons on two sample images from the Video Shadow Removal dataset~\cite{le2020from} with the Mask-ShadowGAN~\cite{hu2019mask} and LG-ShadowNet~\cite{liu2021shadow}.}
    \label{fig:video}
	\vspace{-8pt}
\end{figure}

We also show visualisation comparison results with Mask-ShadowGAN~\cite{hu2019mask} and LG-ShadowNet~\cite{liu2021shadow} on two samples from the video dataset in Fig.~\ref{fig:video}. The shadow regions in our results look lighter with less artefacts than others for images of either close (top) or distant shots (bottom).

\vspace{-4pt}

\section{Conclusion}
To conclude, we proposed a novel G2R-ShadowNet for weakly-supervised shadow removal which is trained without using shadow-free images. The training of the network consists of shadow generation, shadow removal and refinement, which correspond to three sub-nets in G2R-ShadowNet, respectively, and they are jointly trained in an end-to-end fashion. Shadow generation is a prerequisite which stylises non-shadow regions to be shadow ones and constructs paired training set for shadow removal. Extensive experiments showed the effectiveness of our proposed G2R-ShadowNet and verified that our method outperforms the best weakly-supervised method on the adjusted ISTD dataset and the Video Shadow Removal dataset. It also achieved competitive performances against the other state-of-the-arts that use more training data, such as paired or unpaired shadow-free images, than our method.

\paragraph{Acknowledgements.}
This work was partially supported by the Research and Development Program of Beijing Municipal Education Commission (KJZD20191000402) and by the National Nature Science Foundation of China (51827813, 61472029, 61672376, U1803264).

{\small
	\bibliographystyle{ieee_fullname}
	\bibliography{egbib}
}
	
\end{document}